\documentclass[review]{elsarticle}
\graphicspath{ {./figures/} }
\usepackage{hyperref}
\usepackage{float}
\usepackage{verbatim} 
\usepackage{apalike}
\restylefloat{figure}
\restylefloat{table}

\usepackage{algpseudocode}
\usepackage[linesnumbered,ruled,vlined]{algorithm2e}
\usepackage{graphicx} 
\usepackage{grffile}
\usepackage{subcaption}
\usepackage{lineno,hyperref}
\usepackage{amsmath,amssymb,amsfonts}
\usepackage{units}
\usepackage[table,dvipsnames]{xcolor}
\usepackage{tikz}  
\usepackage{tikz-3dplot}
\usepackage{xifthen}
\modulolinenumbers[5]
\tdplotsetmaincoords{60}{125}
\tdplotsetrotatedcoords{0}{0}{0} 
\usetikzlibrary{patterns}
\usetikzlibrary{arrows,shapes,positioning}
\usetikzlibrary{decorations.markings}
\tikzstyle arrowstyle=[scale=1.4]
\tikzstyle directed=[postaction={decorate,decoration={markings,
    mark=at position .65 with {\arrow[arrowstyle]{stealth}}}}]
\usepackage{pgfplots}
\newlength\fwidth

\journal{Expert Systems with Applications}

\biboptions{authoryear}
\DeclareUnicodeCharacter{2217}{*}
\begin{document}
\begin{frontmatter}

\begin{titlepage}
\begin{center}
\vspace*{1cm}

\textbf{ \large  D$^*_+$: A Risk Aware Platform Agnostic Heterogeneous Path Planner}

\vspace{1.5cm}

Samuel Karlsson$^{a}$ (samkar@ltu.se), Anton Koval$^a$ (antkov@ltu.se), Christoforos Kanellakis$^a$ (christoforos.kanellakis@ltu.se), George Nikolakopoulos$^a$ (geonik@ltu.se)\\

\hspace{10pt}

\begin{flushleft}
\small  
$^a$ Robotics \& AI Team, Department of Computer, Electrical and Space Engineering,\\ Lule\r{a} University of Technology, Lule\r{a} SE-97187, Sweden \\


\vspace{1cm}
\textbf{Corresponding Author:} \\
Samuel Karlsson \\
Robotics \& AI Team, Department of Computer, Electrical and Space Engineering,\\ Lule\r{a} University of Technology, Lule\r{a} SE-97187, Sweden \\
Tel: +4670-221 48 33 \\
Email: samkar@ltu.se

\end{flushleft}        
\end{center}
\end{titlepage}

\title{D$^*_+$: A Risk Aware Platform Agnostic Heterogeneous Path Planner}

\author[label1]{Samuel Karlsson \corref{cor1}}
\ead{samkar@ltu.se}

\author[label1]{Anton Koval}
\ead{antkov@ltu.se}

\author[label1]{Christoforos Kanellakis}
\ead{christoforos.kanellakis@ltu.se}

\author[label1]{George Nikolakopoulos}
\ead{geonik@ltu.se}

\cortext[cor1]{Corresponding author.}
\address[label1]{Robotics \& AI Team, Department of Computer, Electrical and Space Engineering,\\ Lule\r{a} University of Technology, Lule\r{a} SE-97187, Sweden \\}
\fntext[fn1]{This work has been partially funded by the European Unions Horizon 2020 Research and Innovation Programme under the Grant Agreement No. 869379 illuMINEation.}

\begin{abstract}
This article establishes the novel D$^*_+$, a risk-aware and platform-agnostic heterogeneous global path planner for robotic navigation in complex environments. The proposed planner addresses a fundamental bottleneck of occupancy-based path planners related to their dependency on accurate and dense maps. More specifically, their performance is highly affected by poorly reconstructed or sparse areas (e.g. holes in the walls or ceilings) leading to faulty generated paths outside the physical boundaries of the 3-dimensional space. As it will be presented, D$^*_+$ addresses this challenge with three novel contributions, integrated into one solution, namely: a) the proximity risk, b) the modeling of the unknown space, and c) the map updates. By adding a risk layer to spaces that are closer to the occupied ones, some holes are filled, and thus the problematic short-cutting through them to the final goal is prevented. The novel established D$^*_+$ also provides safety marginals to the walls and other obstacles, a property that results in paths that do not cut the corners that could potentially disrupt the platform operation. D$^*_+$ has also the capability to model the unknown space as risk-free areas that should keep the paths inside, e.g in a tunnel environment, and thus heavily reducing the risk of larger shortcuts through openings in the walls. D$^*_+$ is also introducing a dynamic map handling capability that continuously updates with the latest information acquired during the map building process, allowing the planner to use constant map growth and resolve cases of planning over outdated sparser map reconstructions. The proposed path planner is also capable to plan 2D and 3D paths by only changing the input map to a 2D or 3D map and it is independent of the dynamics of the robotic platform. The efficiency of the proposed scheme is experimentally evaluated in multiple real-life experiments where D$^*_+$ is producing successfully proper planned paths, either in 2D in the use case of the Boston dynamics Spot robot or 3D paths in the case of an unmanned areal vehicle in varying and challenging scenarios.
\end{abstract}

\begin{keyword}
D$^*_+$, Path planing \sep Risk Aware \sep Platform Agnostic
\end{keyword}

\end{frontmatter}

\section{Introduction}
Robots are becoming more and more common for a large variety of tasks in real-life challenging environments, including but not limited to search and rescue~\cite{agha2021nebula, MISHRA20201,doi:10.1126/scirobotics.abg1188, Hayat2020,SanJuan2018}, inspection~\cite{ZHANG2021103123}, and delivery~\cite{SHE2021102878}. In general, robotic competitions and similar events like the recent DARPA's SubTerranean Challenge~\cite{subtworld} have increased the popularity of robotic platforms, their interaction, and their collaborative operation. With the utilization of multiple robotic platforms, large areas can be covered collaboratively, or multiple tasks can be executed simultaneously with increased performance. However, with the increased number of robots, the complexity increases as well, while the overall usability reduces for every robot, a reality that is caused by the utilization of unique robot-specific software and corresponding settings. One way to simplify the challenge of operating multiple robots is to unify parts of the software independently of the type of platform. Path planning is such an area where a platform-agnostic algorithm could be beneficial and thus in this article we propose a novel D$^*_+$ platform-agnostic global path planner and experimentally evaluate it with the Boston Dynamics (BD) Spot quadruped robot and a quadrotor.

Some path-planning algorithms base their operation on the utilization of environmental characteristics to generate paths in the operational environment. As an example, one can mention the COMPRA framework~\cite{lindqvist2021compra} where the narrow nature of tunnels is utilized to fly toward the deepest point of the tunnel and thus explore it in an efficient way. However, this method works well in tunnels and similar environments, but it is not able to handle junctions and open spaces in a good way.
For similar reasons, using a heterogeneous path planner, like the herein novel proposed D$^*_+$ that is able to work efficiently in many environments, is a most desired need.

Many path planners utilize an occupancy map~\cite{9367235,wang2021learningbased,hornung13auro,ZHANG2021103123} to plan upon, as for example, RRT$^*$, A$^*$, theta$^*$, D$^*$, and the jump point search~\cite{Wu2019,doi:10.2514/6.2018-1846}. However, these planners do rely on a dense occupancy map in order to avoid the problematic situations of generating invalid paths or shortcuts that are planned through obstacles and wall openings. Multiple ways of plugging such holes have been suggested, like filtering~\cite{ATAPOURABARGHOUEI201839} or inflation~\cite{LI2018275} for example. Filtering can solve many imperfections of a map, however, it can also filter out smaller features, such as small actual openings in the operational environment. Furthermore, inflation of the occupied spaces is a simple solution that creates a safety marginal that forces the path to be planned inside the safe region and thus has some marginal for errors. However, inflation is a hard artificially created boundary that can block a path, where a robot would have been able to go, even though it would be a higher risk passage. Alike these two approaches, risk-aware-based path planners can accept some risks to reach the goal without taking unnecessary risks~\cite{laconte2021novel}.

Utilizing a risk assessment to plan for a better path is something that has been addressed previously in the scientific literature as in~\cite{9483405,9341084}. However, one of the biggest challenges, when working with risk maps, is how to generate the risk and what to consider as a risk. In many realistic application scenarios, a type of risk that can be considered is the consequences if a crash or failure occurs~\cite{9165709, 7991358}, and then minimizing the consequences of a crash. An alternative method is to utilize offline and online risk maps to plan crash-safe paths as in~\cite{Primatesta2019,da2019collision}, while this approach is considering a 2D risk map in open areas to plan their paths. Considering the consequences of a crash is a very good approach to enable good paths, however, it does not aid in actually avoiding a true collision, while for the case of path planning for Unmanned Aerial Vehicles (UAVs) that can move in 3D, the risk assessment and path planning algorithms should utilize all three dimensions when planning a path~\cite{hakobyan2019risk}. Risk considerations for safe path planning in a confined environment and by using 3D maneuvering is barely done in the literature. Thus, one of the few works as in~\cite{zhou2021raptor} utilizes a prediction risk of unknown areas and controls the yaw angle to identify and calculate potential dangers as early as possible. For the case of ground robots, rough and uneven surfaces can be considered as a realistic risk~\cite{7119022, puck2020modular}, however, sharp rocks and loose sand that are true risks for ground robots do not constitute real relevant risks for UAVs unless it is so loose that the UAV creates dust out of it. As such, depending on the environment and the utilized robotic platform, different things can be identified as a risk and with a different estimation of their severity impact, while one of the few things that are always dangerous for the robotic operation is the existence of obstacles and the corresponding proximity to the obstacles.

Many path planners try to plan the shortest possible path, either by ensuring that it as a grid search method like the D$^*$ style planners or as sample-based planners, like the RRT$^*$ that performs pseudo-random guessing of a path and settles for one that is good enough. In this approach, the shortest path around a corner is tangent to the inside corner, but although this is a free space, it is not necessarily safe for a robot to be that close to an obstacle, mainly due to inaccuracies in mapping, state estimation, and path tracking. Thus, one option to generate safe paths is to use model constraints, such in the methods in~\cite{Yan2013,tordesillas2019faster,tordesillas2021faster} when planning. These path planners are able to generate paths with the knowledge of robot kinematics and ensure that the path is possible to traverse. One of the main disadvantages is the complexity of these planners is the need to know the kinematic model of the robot, which makes it hard to allow for a generalized application of these algorithms between different robots.

Among the platform agnostic planners, one can highlight the Graph-based Exploration Planner 2.0 (Gbplanner)~\cite{9812401,dang2020graph}, which was developed and utilized by team CERBERUS who won the DARPA subterranean challenge~\cite{subtworld}. Gbplanner is a combined exploration and path planning solution that builds a graph through a RRT local planning step. After the graph is built, Dijkstra's algorithm is used to rapidly plan global paths on it. In our work, we will evaluate our planner against Gbplanner, which is a state-of-the-art path planning solution.

Thus, the main contribution of this article is the establishment of the novel D$^*_+$ occupancy-based risk-aware, platform-agnostic, heterogeneous global path planner. The novelty of D$^*_+$ stems from the evolution of D$^*$-lite by 1) treating proximity to occupied spaces as risky areas, 2) modeling explicitly unknown areas as a risk, and 3) allowing dynamic 3D map updates for planning. Moreover, D$^*_+$ is a field-hardened global planner that has been extensively tested and evaluated on a quadruped Spot robot and an UAV. D$^*_+$ stands, to the best of the author's knowledge, one of the few global path planners that are able to avoid obstacles by maneuvering in full 3D and considering risks while doing so. At the same time, D$^*_+$ is the first algorithm where these types of risks and path planning methods are used in combination. Finally, the code has also been made open source for the robotics community and is available on GitHub~\footnote{Link to the code on GitHub: \url{https://github.com/LTU-RAI/Dsp}}. 

\subsection{Outline}
The rest of the article is structured as follows. In Section~\ref{sec:dsp} a complete presentation of the architecture and core components of D$^*_+$ is provided, while in Section~\ref{sec:exp} the experimental setups for the algorithmic evaluation are explained. In Section~\ref{sec:result} the extended experimental evaluation of the suggested scheme is presented, while future work and conclusions are drawn in section~\ref{sec:con}.
%
\section{D$^*_+$}\label{sec:dsp}
%
The D$^*$-lite~\cite{koenig2002d} algorithm is able to plan a path on a grid ($\mathbb{G}$) where it finds the shortest path $P_{\mathbf{A} \to \mathbf{B}}$ from the point $\mathbf{A}$ to the point $\mathbf{B}$, based on the traversal cost ($C$) of the voxels ($v$) in $\mathbb{G}$ so that the $\sum{\forall C \in P_{\mathbf{A} \to \mathbf{B}}}$ is the smallest possible cost, while another benefit of D$^*$-lite is the way it saves previous path calculations to speed up the re-planning of $P$.

The main difference between D$^*$-lite and D$^*_+$ is how $\mathbb{G}$ is constructed from an input map $\mathbb{M}$. In the proposed scheme, D$^*_+$ is creating $\mathbb{G}$ with the consideration of free, occupied and unknown voxels ($v_f,v_o,v_u$) and a risk layer. Since the dimensions of $\mathbb{G}$ are changeable, the D$^*_+$ is also able to plan paths in both 2D and 3D. Thus, by creating $\mathbb{G}$ with only one layer in the $z$-axis it will enable D$^*_+$ to plan a 2D path, while if $\mathbb{G}$ is created with multiple layers in the $z$-axis it will enable for a 3D path. Because D$^*_+$ is planning $P$ within $\mathbb{G}$, any path in Figure~\ref{fig:2dv3d:a} will be a 2D path and any path in Figure~\ref{fig:2dv3d:b} will be a 3D path. Thus, is it possible to use the same algorithm to plan both 2D and 3D paths, where the difference is the map $\mathbb{M}$ from which $\mathbb{G}$ is created.

The proposed D$^*_+$ path planer utilizes three main differences from D$^*$-lite (as presented in~\cite{koenig2002d}) namely: 1) the treatment of unknown voxels, 2) proximity risk and 3) map updates. The rest of this Section describes the proposed algorithm in detail, which can be also summarized in algorithms~\ref{algo:1} -~\ref{algo:fprox}. In Figure~\ref{fig:dspArt} a block diagram illustrating how the algorithms in D$^*_+$ are connected with each other is presented, while it should be mentioned that all the implementations are made in C++ within the Robotic Operating System~\cite{ros} and are open sourced as indicated before.

\begin{figure}
    \centering
    \begin{subfigure}[b]{0.45\textwidth}
        \centering
        \begin{tikzpicture}[scale=0.9,tdplot_rotated_coords,
                    cube/.style={very thick,black},
                    grid/.style={very thin,gray},
                    axis/.style={->,blue,ultra thick},
                    rotated axis/.style={->,purple,ultra thick}]

aw[cube,fill=green!5] (0,0,0) -- (3,0,0) -- (3,0,3) -- (0,0,3) -- cycle;

            \draw[cube,fill=darkgray, opacity=1.0] (1,1,1) -- (3,1,1) -- (3,1,0) -- (1,1,0) -- cycle;
            \draw[cube,fill=darkgray, opacity=1.0] (3,2,1) -- (4,2,1) -- (4,2,0) -- (3,2,0) -- cycle;
            \draw[cube,fill=darkgray, opacity=1.0] (0,4,1) -- (4,4,1) -- (4,4,0) -- (0,4,0) -- cycle;
            
            \draw[cube,fill=lightgray, opacity=1.0] (4,0,1) -- (4,2,1) -- (4,2,0) -- (4,0,0) -- cycle;
            \draw[cube,fill=lightgray, opacity=1.0] (1,1,1) -- (1,3,1) -- (1,3,0) -- (1,1,0) -- cycle;
            \draw[cube,fill=lightgray, opacity=1.0] (4,3,1) -- (4,4,1) -- (4,4,0) -- (4,3,0) -- cycle;
           
            \draw[cube,fill=gray, opacity=0.9] (4,0,1) -- (4,2,1) -- (3,2,1) -- (3,0,1) -- cycle;
            \draw[cube,fill=gray, opacity=0.9] (4,0,1) -- (4,1,1) -- (0,1,1) -- (0,0,1) -- cycle;
            \draw[cube,fill=gray, opacity=0.9] (1,1,1) -- (1,4,1) -- (0,4,1) -- (0,1,1) -- cycle;
            \draw[cube,fill=gray, opacity=0.9] (0,3,1) -- (4,3,1) -- (4,4,1) -- (0,4,1) -- cycle;
    
            \foreach \x in {0,1,2,3,4}
               \foreach \y in {0,1,2,3,4}
                  \foreach \z in {0,1}{
                       \ifthenelse{  \lengthtest{\x pt < 4pt}  }
                       {
                            \draw [black]   (\x,\y,\z) -- (\x+1,\y,\z);
                       }
                       {
                       }
                       \ifthenelse{  \lengthtest{\y pt < 4pt}  }
                       {
                            \draw [black]   (\x,\y,\z) -- (\x,\y+1,\z);
                       }
                       {
                       }
                       \ifthenelse{  \lengthtest{\z pt < 1pt}  }
                       {
                            \draw [black]   (\x,\y,\z) -- (\x,\y,\z+1);
                       }
                       {
                       }
            }

        \end{tikzpicture}
        \caption{}
        \label{fig:2dv3d:a}
    \end{subfigure}
    \qquad
    \begin{subfigure}[b]{0.45\linewidth}
        \centering
        \begin{tikzpicture}[scale=0.9,tdplot_rotated_coords,
                    cube/.style={very thick,black},
                    grid/.style={very thin,gray},
                    axis/.style={->,blue,ultra thick},
                    rotated axis/.style={->,purple,ultra thick}]

aw[cube,fill=green!5] (0,0,0) -- (3,0,0) -- (3,0,3) -- (0,0,3) -- cycle;
 
            \draw[cube,fill=darkgray] (1,1,3) -- (3,1,3) -- (3,1,0) -- (1,1,0) -- cycle;
            \draw[cube,fill=lightgray] (4,0,3) -- (4,2,3) -- (4,2,0) -- (4,0,0) -- cycle;
            \draw[cube,fill=lightgray] (1,1,3) -- (1,3,3) -- (1,3,0) -- (1,1,0) -- cycle;
            \draw[cube,fill=lightgray] (4,3,3) -- (4,4,3) -- (4,4,0) -- (4,3,0) -- cycle;
           
            \draw[cube,fill=darkgray] (3,2,3) -- (4,2,3) -- (4,2,0) -- (3,2,0) -- cycle;
            \draw[cube,fill=darkgray] (0,4,3) -- (4,4,3) -- (4,4,0) -- (0,4,0) -- cycle;
            
            \draw[cube,fill=gray] (4,0,3) -- (4,2,3) -- (3,2,3) -- (3,0,3) -- cycle;
            \draw[cube,fill=gray] (4,0,3) -- (4,1,3) -- (0,1,3) -- (0,0,3) -- cycle;
            \draw[cube,fill=gray] (1,1,3) -- (1,4,3) -- (0,4,3) -- (0,1,3) -- cycle;
            \draw[cube,fill=gray] (0,3,3) -- (4,3,3) -- (4,4,3) -- (0,4,3) -- cycle;
   
            \foreach \x in {0,1,2,3,4}
               \foreach \y in {0,1,2,3,4}
                  \foreach \z in {0,1,2,3}{
                       \ifthenelse{  \lengthtest{\x pt < 4pt}  }
                       {
                            \draw [black]   (\x,\y,\z) -- (\x+1,\y,\z);
                       }
                       {
                       }
                       \ifthenelse{  \lengthtest{\y pt < 4pt}  }
                       {
                            \draw [black]   (\x,\y,\z) -- (\x,\y+1,\z);
                       }
                       {
                       }
                       \ifthenelse{  \lengthtest{\z pt < 3pt}  }
                       {
                            \draw [black]   (\x,\y,\z) -- (\x,\y,\z+1);
                       }
                       {
                       }
            }

        \end{tikzpicture}
        \caption{}
        \label{fig:2dv3d:b}
    \end{subfigure}
    \caption{$\mathbb{G}$ with one $z$-layer from a $\mathbb{M}_2$ and $\mathbb{G}$ with multiple $z$-layers from a $\mathbb{M}_3$. Both $\mathbb{M}_2$ and $\mathbb{M}_3$ are created from the same environment.}
    \label{fig:2dv3d}
\end{figure}
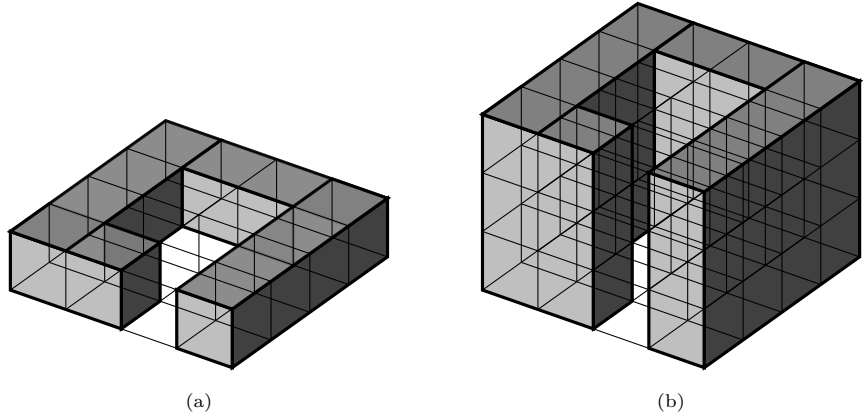
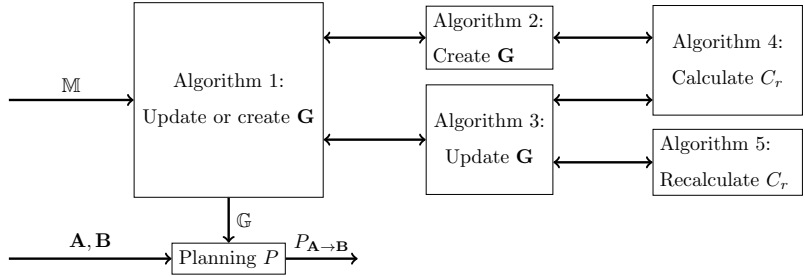
\begin{figure}
\centering
\resizebox{0.9\linewidth}{!}{
\begin{tikzpicture}[scale=1.0]

    \node[draw] at (3.5, 0.7) (lio) [text width=2.4cm, text height=1.7cm,align=center]{}; 
    \node[align=center] at (lio.center) {Algorithm~\ref{algo:prox}:\\Calculate $C_r$};
    \node[draw] at (3.5, -1.1) (octo) [text width=2.4cm]{Algorithm~\ref{algo:fprox}: \\Recalculate $C_r$}; 
    \node[draw] at (-5.3,0.0)(dsp) [text width=3.1cm, text height=3.2cm]{}; 
    \node[align=center] at (dsp.center) {Algorithm~\ref{algo:1}: \\Update or create $\mathbf{G}$};

    \node[draw] at (-0.7,-0.7) (vlp) [text width=2cm, text height=1.7cm,align=center]{}; 
      \node[align=center] at (vlp.center) {Algorithm~\ref{algo:update}:\\ Update $\mathbf{G}$};
    \node[draw] at (-0.7,1.1) (vn) [text width=2cm]{Algorithm~\ref{algo:create}: \\Create $\mathbf{G}$}; 
    
    \node[draw] at (-5.3,-2.8) (dsl) {Planning $P$};
    
    \node at (-9.3,0.0) (m) {};
    \node at (-9.3,-2.8) (o) {};
    \node at (-2.9,-2.8) (p) {};
    
    \node at (-3.75,1.1) (a) {};
    \node at (-3.75,-0.7) (b) {};
    \node at (2.3,1.1) (c) {};
    \node at (2.3,0.0) (d) {};
    \node at (0.3,0.0) (e) {};
    \node at (0.3,-1.1) (f) {};

    \draw[<->,very thick] (e) edge node[] {} (d);
    \draw[<->,very thick] (a) edge node[] {} (vn);
    \draw[<->,very thick] (b) edge node[] {} (vlp);
    \draw[<->,very thick] (vn) edge node[] {} (c);
    \draw[<->,very thick] (f) edge node[] {} (octo);
    \draw[->,very thick] (dsp) edge node[right] {$\mathbb{G}$} (dsl);
    \draw[->,very thick] (m) edge node[above] {$\mathbb{M}$} (dsp);
    \draw[->,very thick] (o) edge node[above] {$\mathbf{A},\mathbf{B}$} (dsl);
    \draw[->,very thick] (dsl) edge node[above] {$P_{\mathbf{A} \to \mathbf{B}}$} (p);

\end{tikzpicture}
}
\caption{A block diagram illustrating how the algorithmic blocks in D$^*_+$ are connected. Algorithm~\ref{algo:1} receives updated $\mathbb{M}$ and decides if $\mathbb{G}$ should be created from a scratch (algorithm~\ref{algo:create}) or updated (algorithm~\ref{algo:update}). Algorithms~\ref{algo:prox} -~\ref{algo:fprox} are used as help functions for algorithms~\ref{algo:create} -~\ref{algo:update} to generate risk for $\mathbb{G}$.}
\label{fig:dspArt}
\end{figure}
\subsection{Modeling of Unknown Space}
\begin{figure}
    \centering
    \includegraphics[width=0.8\textwidth]{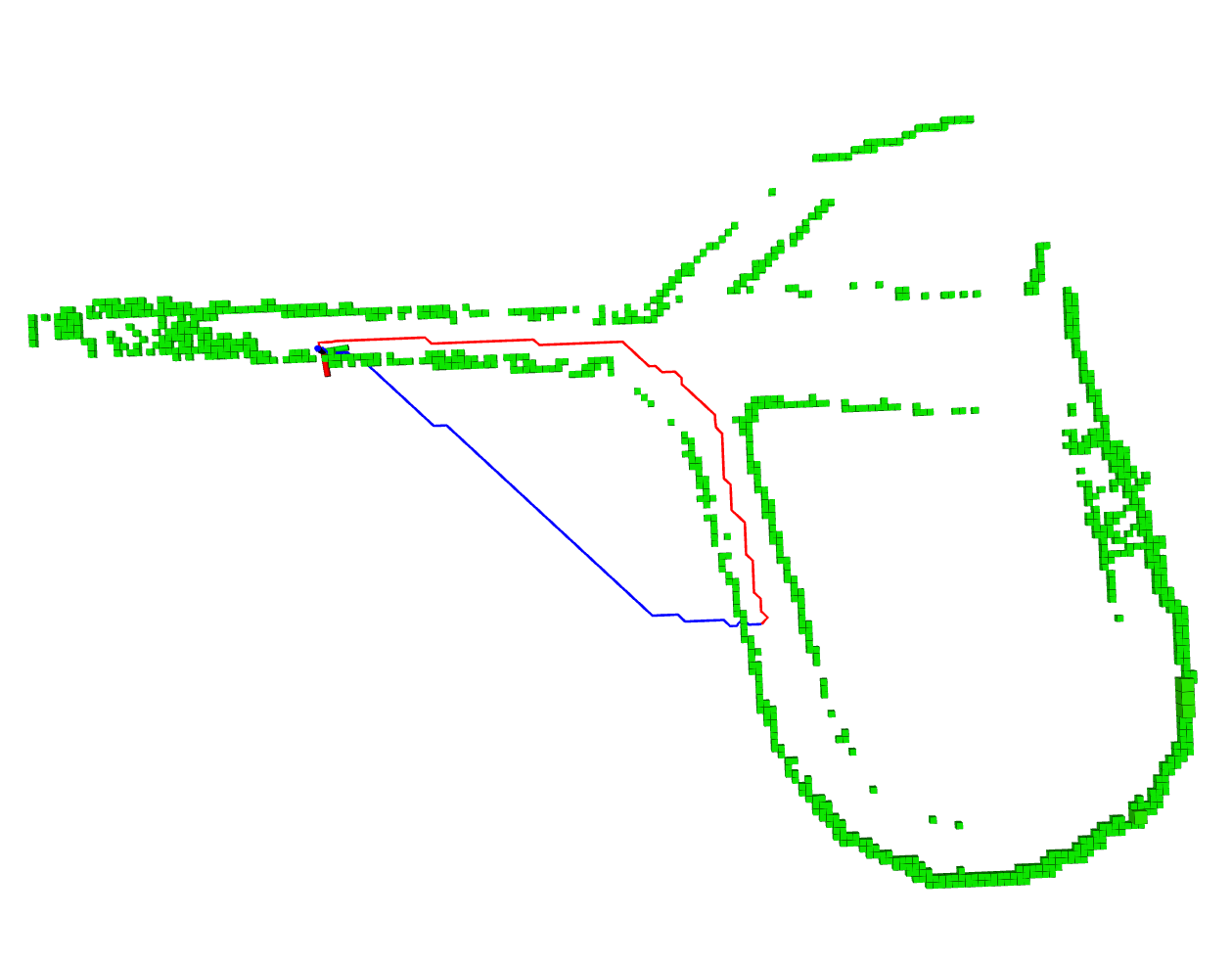}
    \caption{A comparison of path planning with (red line) and without (blue line) consideration of $v_u$ under the same input $\mathbb{M}$, starting and endpoint. This comparison does also show the difference between D$^*_+$ and D$^*$-lite (adjusted for online usage with an octomap). With the utilization of $v_u$ the planned path remains in the caving environment, while without the path it becomes invalid while passing through the solid rock wall.}
    \label{fig:represent}
\end{figure}

Imperfections in $\mathbb{M}$ will lead to imperfections in $\mathbb{G}$ that may lead to path short-cutting through walls like the blue path in Figure~\ref{fig:represent}. In D$^*_+$ this problem is solved by having $v_u$ represented in $\mathbb{G}$, as well as $v_f$ and $v_o$. 
Let $C_f, C_u, C_o$ costs assigned to $v_f, v_u, v_o$ respectively so that $C_f < C_u < C_o$, where in such case $P$ will be planned through $v_f$ instead of $v_u$ even though the resulting $P$ length is longer. This results from the core of D$^*_+$ that plans the path where $\sum{\forall C \in P_{\mathbf{A} \to \mathbf{B}}}$ is the smallest possible.
With a higher $C_u$ value the D$^*_+$ will plan the $P$ in $v_f$ at a cost of longer $P$'s than it would make with a smaller $C_u$.

\subsection{Path planning with a proximity risk}

\begin{figure}[ht]
    \centering
    \includegraphics[width=0.9\textwidth]{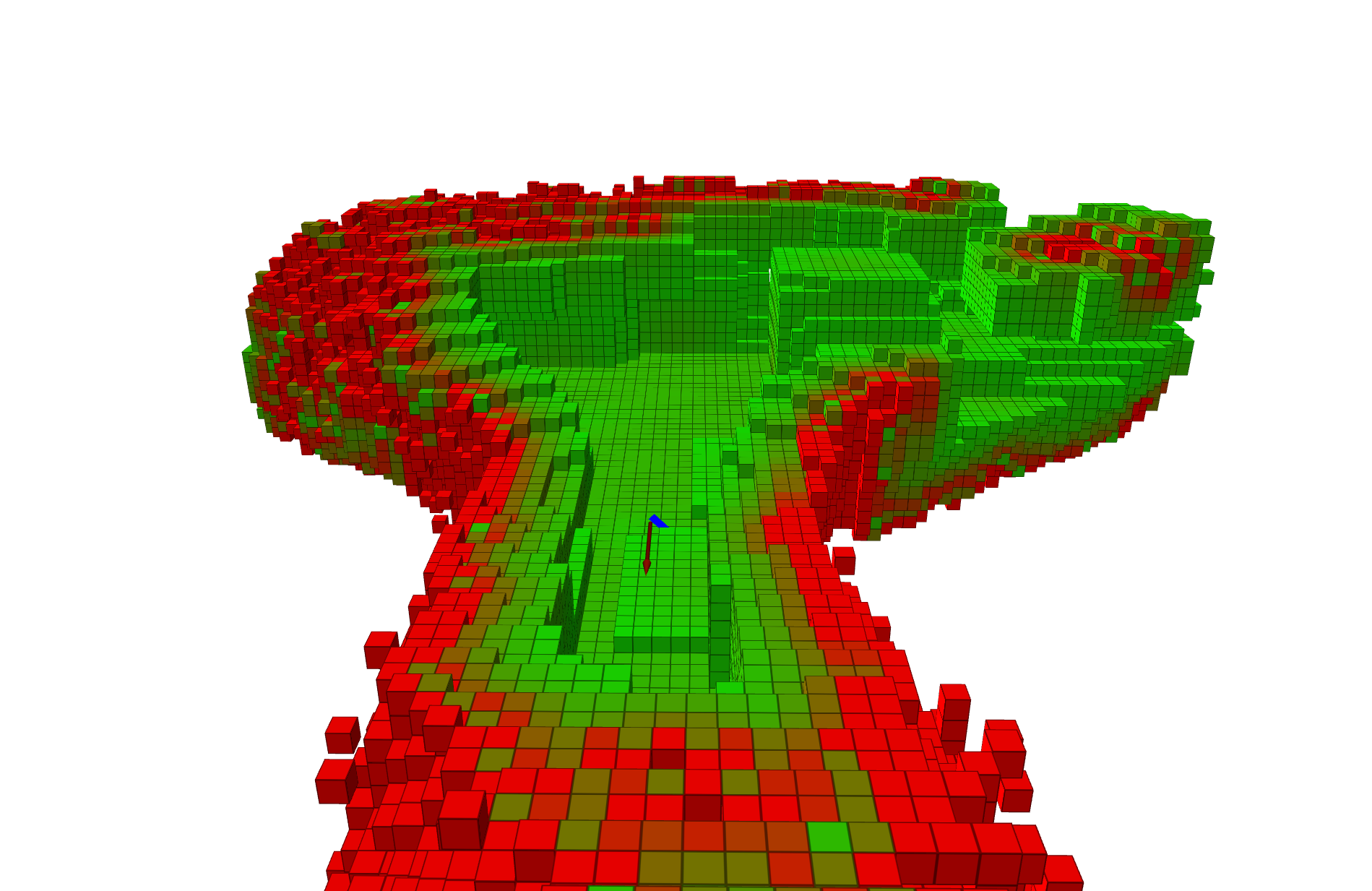}
    \caption{A visualization of $C_r$, where red is a high risk and green is a low risk. It should be noted that $V_f$ and $v$ with $C \geq C_u$ are not visualized.}
    \label{fig:gradientRis}
\end{figure}

D$^*$ style path planners are created to plan the shortest path from point $\mathbf{A}$ to $\mathbf{B}$, and the shortest path around a bend, or corner is to hug, meaning traverse as close as possible inside of the corner of interest. As such, the utilization of such planners introduces the issue that the planned path leaves no room for errors in path following and is even setting limitations to the robot's physical size. To deal with this problematic situation, in D$^*_+$ a risk layer is introduced, where $v$'s in the proximity of a $v_o$ are given an extra traversal cost $C_r$ and as such the traversal cost of a $v$ in the proximity of a $v_o$ can be defined as:

\begin{align}
    C_r &= \begin{cases}
        \frac{C_u}{d + 1} & \text{if $d < r$} \\
        0 & \text{else} \\
    \end{cases}\\
    C &= \begin{cases}
        C_f + C_r & \text{if $v$ is $v_f$} \\
        C_u + C_r & \text{if $v$ is $v_u$}
    \end{cases}\\
\end{align}
where $d$ is the distance in voxels to the closest $v_o$ and $r$ is the range that is considered as the proximity to a $v_o$. This novel approach creates a gradient risk that has the highest value next to $v_o$ and decreases with distance as depicted in Figure~\ref{fig:gradientRis} where the $v$ with a $C_r$ is depicted.

\begin{figure}
    \centering

\begin{tikzpicture}[scale=1,tdplot_rotated_coords,
                    cube/.style={very thick,black},
                    grid/.style={very thin,gray},
                    axis/.style={->,blue,ultra thick},
                    rotated axis/.style={->,purple,ultra thick}]

aw[cube,fill=green!5] (0,0,0) -- (3,0,0) -- (3,0,3) -- (0,0,3) -- cycle;
    

\foreach \x in {0,1,2,3,4,5,6,7}
   \foreach \y in {0,1,2,3,4,5,6,7,8,9}
      \foreach \z in {0,1,2}{
           \ifthenelse{  \lengthtest{\x pt < 7pt}  }
           {
                \draw [black]   (\x,\y,\z) -- (\x+1,\y,\z);
           }
           {
           }
           \ifthenelse{  \lengthtest{\y pt < 9pt}  }
           {
                \draw [black]   (\x,\y,\z) -- (\x,\y+1,\z);
           }
           {
           }
           \ifthenelse{  \lengthtest{\z pt < 2pt}  }
           {
                \draw [black]   (\x,\y,\z) -- (\x,\y,\z+1);
           }
           {
           }
}    
    
    \shade[ball color=red!70!white] (0.5,0.5,1.5) circle (.2);

     \draw[thick, directed] (2.5, 0.5, 1.5) -- (0.5, 0.5, 1.5);
     
 \definecolor{high}{rgb}{1,0.3,0.3}
 \definecolor{mid}{rgb}{0.7,0.7,0.3}
     \definecolor{low}{rgb}{0.3,1.0,0.3}
    \draw[cube,fill=darkgray, opacity=0.9, draw=none] (0,4,0) -- (1,4,0) -- (1,4,2) -- (0,4,2) -- cycle;
    \draw[cube,fill=darkgray, opacity=0.9, draw=none] (1,3,0) -- (1,4,0) -- (1,4,2) -- (1,3,2) -- cycle;
    
    \draw[cube,fill=high, opacity=0.6, draw=none] (0,5,2) -- (0,5,0) -- (2,5,0) -- (2,5,2) -- cycle;
    \draw[cube,fill=high, opacity=0.6, draw=none] (2,2,0) -- (2,5,0) -- (2,5,2) -- (2,2,2) -- cycle;
    
    \draw[cube,fill=low, opacity=0.6, draw=none] (0,1,2) -- (0,6,2) -- (3,6,2) -- (3,1,2) -- cycle;
    \draw[cube,fill=high, opacity=0.6, draw=none] (0,2,2) -- (0,5,2) -- (2,5,2) -- (2,2,2) -- cycle;
    \draw[cube,fill=darkgray, opacity=0.9, draw=none] (0,3,2) -- (0,4,2) -- (1,4,2) -- (1,3,2) -- cycle;
    
    \draw[cube,fill=low, opacity=0.6, draw=none] (0,6,2) -- (0,6,0) -- (3,6,0) -- (3,6,2) -- cycle;
    \draw[cube,fill=low, opacity=0.6, draw=none] (3,1,0) -- (3,6,0) -- (3,6,2) -- (3,1,2) -- cycle;

    \draw[thick, directed] (3.5, 7.5, 1.5) -- (2.5, 6.5, 1.5);
     \draw[thick, directed] (2.5, 6.5, 1.5) -- (2.5, 2.5, 1.5);
     \draw[thick, directed] (2.5, 2.5, 1.5) -- (3.5, 1.5, 1.5);
     \draw[thick, directed] (3.5, 1.5, 1.5) -- (2.5, 0.5, 1.5);
    
    \draw[cube,fill=darkgray, opacity=0.9, draw=none] (6,5,1) -- (6,6,1) -- (5,6,1) -- (5,5,1) -- cycle;
    \draw[cube,fill=darkgray, opacity=0.9, draw=none] (6,5,1) -- (5,5,1) -- (5,5,2) -- (6,5,2) -- cycle;
    \draw[cube,fill=darkgray, opacity=0.9, draw=none] (6,6,0) -- (5,6,0) -- (5,6,1) -- (6,6,1) -- cycle;
    \draw[cube,fill=darkgray, opacity=0.9, draw=none] (7,6,0) -- (6,6,0) -- (6,6,2) -- (7,6,2) -- cycle;
    \draw[cube,fill=darkgray, opacity=0.9, draw=none] (7,0,2) -- (7,6,2) -- (6,6,2) -- (6,0,2) -- cycle;
    \draw[cube,fill=darkgray, opacity=0.9, draw=none] (7,0,0) -- (7,6,0) -- (7,6,2) -- (7,0,2) -- cycle;
    \draw[cube,fill=darkgray, opacity=0.9, draw=none] (6,4,2) -- (6,5,2) -- (5,5,2) -- (5,4,2) -- cycle;
    
    \draw[cube,fill=high, opacity=0.6, draw=none] (6,0,2) -- (6,4,2) -- (5,4,2) -- (5,0,2) -- cycle;
    \draw[cube,fill=high, opacity=0.6, draw=none] (5,3,2) -- (5,6,2) -- (4,6,2) -- (4,3,2) -- cycle;
    \draw[cube,fill=high, opacity=0.6, draw=none] (7,6,2) -- (7,7,2) -- (5,7,2) -- (5,6,2) -- cycle;
    \draw[cube,fill=high, opacity=0.6, draw=none] (6,5,2) -- (6,6,2) -- (5,6,2) -- (5,5,2) -- cycle;
    \draw[cube,fill=high, opacity=0.6, draw=none] (5,6,1) -- (5,7,1) -- (4,7,1) -- (4,6,1) -- cycle;
    
    \draw[cube,fill=high, opacity=0.6, draw=none] (5,6,0) -- (4,6,0) -- (4,6,2) -- (5,6,2) -- cycle;
    \draw[cube,fill=high, opacity=0.6, draw=none] (7,7,0) -- (5,7,0) -- (5,7,2) -- (7,7,2) -- cycle;
    \draw[cube,fill=high, opacity=0.6, draw=none] (5,7,0) -- (4,7,0) -- (4,7,1) -- (5,7,1) -- cycle;
    \draw[cube,fill=high, opacity=0.6, draw=none] (7,6,0) -- (7,7,0) -- (7,7,2) -- (7,6,2) -- cycle;

    \draw[cube,fill=low, opacity=0.6, draw=none] (5,0,2) -- (5,3,2) -- (4,3,2) -- (4,0,2) -- cycle;
    \draw[cube,fill=low, opacity=0.6, draw=none] (4,2,2) -- (4,7,2) -- (3,7,2) -- (3,2,2) -- cycle;
    \draw[cube,fill=low, opacity=0.6, draw=none] (7,7,2) -- (7,8,2) -- (4,8,2) -- (4,7,2) -- cycle;
    \draw[cube,fill=low, opacity=0.6, draw=none] (5,6,2) -- (5,7,2) -- (4,7,2) -- (4,6,2) -- cycle;
    \draw[cube,fill=low, opacity=0.6, draw=none] (4,7,1) -- (4,8,1) -- (3,8,1) -- (3,7,1) -- cycle;
    \draw[cube,fill=low, opacity=0.6, draw=none] (4,8,0) -- (3,8,0) -- (3,8,1) -- (4,8,1) -- cycle; 
    
    \draw[thick, directed] (4.5, 8.5, 0.5) -- (3.5, 7.5, 1.5);
    
    \draw[cube,fill=low, opacity=0.6 , draw=none] (4,7,1) -- (3,7,1) -- (3,7,2) -- (4,7,2) -- cycle;
    \draw[cube,fill=low, opacity=0.6, draw=none] (7,8,0) -- (4,8,0) -- (4,8,2) -- (7,8,2) -- cycle;

    \draw[cube,fill=low, opacity=0.6, draw=none] (7,7,0) -- (7,8,0) -- (7,8,2) -- (7,7,2) -- cycle;
    
    \draw [black]   (0,1,2) -- (3,1,2);
    \draw [black]   (3,1,2) -- (3,6,2);
    \draw [black]   (3,6,2) -- (0,6,2);
    \draw [black]   (0,1,2) -- (0,6,2);
    \draw [black]   (0,2,2) -- (2,2,2);
    \draw [black]   (2,2,2) -- (2,5,2);
    \draw [black]   (2,5,2) -- (0,5,2);
    
    \draw [black]   (0,3,2) -- (1,3,2);
    \draw [black]   (1,3,2) -- (1,4,2);
    \draw [black]   (1,4,2) -- (0,4,2);
    
    \draw [black]   (0,6,2) -- (0,6,0);
    \draw [black]   (3,6,0) -- (0,6,0);

    \draw[black] (4,0,2) -- (4,2,2);
    \draw[black] (4,2,2) -- (3,2,2);
    \draw[black] (3,2,2) -- (3,7,2);
    \draw[black] (3,7,2) -- (4,7,2);
    \draw[black] (4,7,2) -- (4,8,2);
    \draw[black] (4,8,2) -- (7,8,2);
    
    \draw[black] (3,8,1) -- (3,7.3,1);
    \draw[black] (4,8,1) -- (3,8,1);
    \draw[black] (4,8,2) -- (4,8,1);
    \draw[black] (3,7,2) -- (3,7,1.3);
    
    \draw[black] (4,0,2) -- (7,0,2);
    \draw[black] (7,0,2) -- (7,8,2);
    \draw[black] (5,0,2) -- (5,3,2);
    \draw[black] (5,3,2) -- (4,3,2);
    \draw[black] (4,3,2) -- (4,6,2);
    \draw[black] (4,6,2) -- (5,6,2);
    \draw[black] (5,6,2) -- (5,7,2);
    \draw[black] (5,7,2) -- (7,7,2);
    
    \draw[black] (6,0,2) -- (6,4,2);
    \draw[black] (6,4,2) -- (5,4,2);
    \draw[black] (5,4,2) -- (5,5,2);
    \draw[black] (5,5,2) -- (6,5,2);
    \draw[black] (6,5,2) -- (6,6,2);
    \draw[black] (6,6,2) -- (7,6,2);
    
    \draw[black] (7,0,2) -- (7,0,0);
    \draw[black] (7,6,2) -- (7,6,0);
    \draw[black] (7,7,2) -- (7,7,0);
    \draw[black] (7,8,2) -- (7,8,0);
    \draw[black] (3,8,1) -- (3,8,0);
    
    \draw[black] (3,8,0) -- (7,8,0);
    \draw[black] (7,8,0) -- (7,0,0);
    
    \draw[black] (1,7,2) -- (3,7,2);
    \draw[black] (1,6,2) -- (1,7,2);
    \draw[black] (2,6,2) -- (2,7,2);
    \draw[black] (2,8,2) -- (2,7,2);
    \draw[black] (3,8,2) -- (3,7,2);
    \draw[black] (2,6.3,1) -- (2,7,1);
    \draw[black] (1,6,1) -- (1,7,1);
    \draw[black] (2,7,2) -- (2,7,1);
    
    \draw[black] (5,9,2) -- (7,9,2);
    \draw[black] (5,8,2) -- (5,9,2);
    \draw[black] (6,8,2) -- (6,9,2);
    \draw[black] (7,8,2) -- (7,9,2);
    \draw[black] (4,8,1) -- (4,9,1);
    \draw[black] (5,8,1) -- (5,9,1);
    \draw[black] (6,8,1) -- (6,9,1);
    \draw[black] (7,8,1) -- (7,9,1);
    \draw[black] (5,9,2) -- (5,9,1);
    \draw[black] (6,9,2) -- (6,9,1);
    \draw[black] (7,9,2) -- (7,9,1);

    
\shade[ball color=blue!70!white] (6.5,8.5,0.5) circle (.2);
\draw[thick, directed] (6.5, 8.5, 0.5) -- (4.5, 8.5, 0.5);
\end{tikzpicture}

        \caption{An illustration of how D$^*_+$ plans a path from the blue ball to the red ball with regards to the risk layer, red voxels are higher risk, green voxels are lower risk, and black voxels are occupied.}
        \label{fig:riskViz}
    \end{figure}
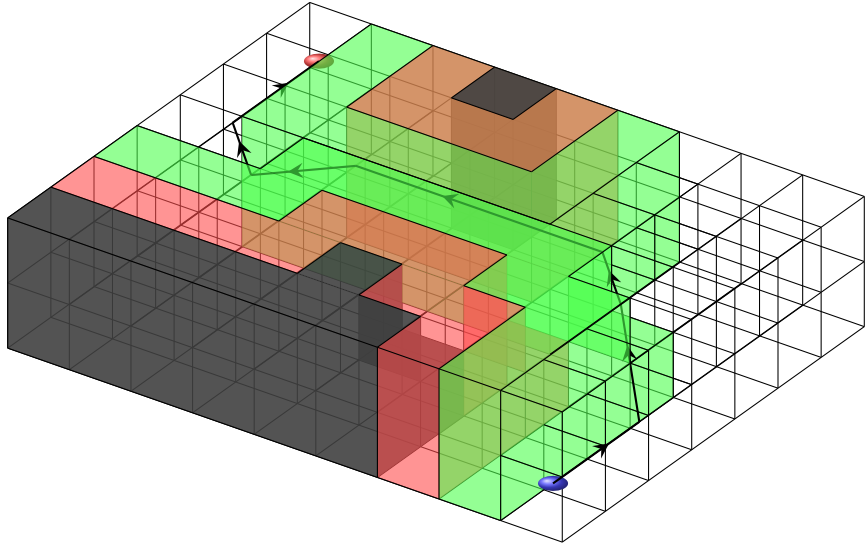

A direct effect of the proposed D$^*_+$ approach is that in places with sufficient space, this risk layer will enable the planning of a path that is outside of the risk area. However, in the cases where a passage is narrow, the gradient of the risk will cause the path to be planned in the center of the passage, as illustrated in Figure~\ref{fig:riskViz}. As such, this novel addition allows D$^*_+$ to plan paths in narrow passages and have a larger safety margin in more open areas, especially when compared with similar path planners.

\subsection{Continuous map updates and expansions during mission}
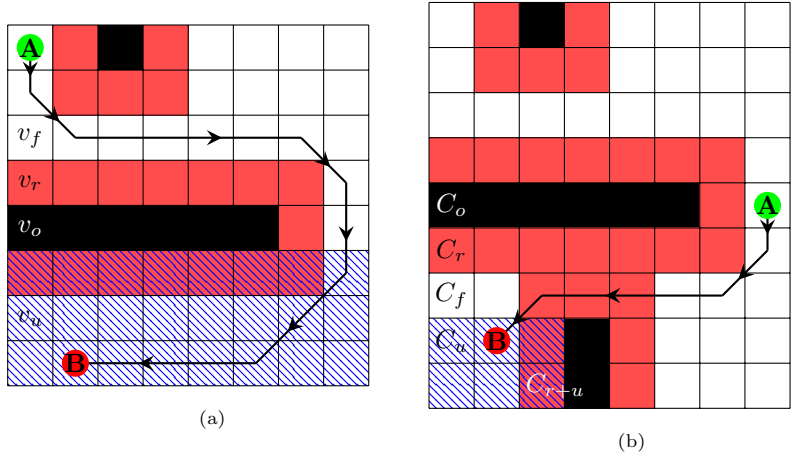
\begin{figure}
    \centering
\begin{subfigure}{0.45\linewidth}
\begin{tikzpicture}[scale=0.6]
    \definecolor{high}{rgb}{1,0.3,0.3}
    
    \fill[black] (2,8) rectangle (3,7);
    \fill[black] (0,4) rectangle (6,3);
    
    \fill[high] (1,8) rectangle (2,6);
    \fill[high] (3,8) rectangle (4,6);
    \fill[high] (1,7) rectangle (4,6);
    
    \fill[high] (0,5) rectangle (6,4);
    \fill[high] (6,5) rectangle (7,2);
    \fill[high] (0,3) rectangle (7,2);
    
    
    \fill [pattern = north west lines, pattern color=blue] (0,3) rectangle (8,0);
    
    \draw (0,0) grid (8,8);

    \draw[thick, directed] (0.5, 7.5) -- (0.5,6.5);
    \draw[thick, directed] (0.5, 6.5) -- (1.5,5.5);
    \draw[thick, directed] (1.5, 5.5) -- (6.5,5.5);
    \draw[thick, directed] (6.5, 5.5) -- (7.5,4.5);
    \draw[thick, directed] (7.5, 4.5) -- (7.5,2.5);
    \draw[thick, directed] (7.5, 2.5) -- (5.5,0.5);
    \draw[thick, directed] (5.5, 0.5) -- (1.5,0.5);
   
    \fill[green] (0.5,7.5) circle (0.3);
    \fill[red] (1.5,0.5) circle (0.3);
    \node[] at (0.5,7.5) {$\mathbf{A}$};
    \node[] at (1.5,0.5) {$\mathbf{B}$};
    
    \node[white] at (0.5,3.5) {$v_o$};
    \node[] at (0.5,5.5) {$v_f$};
    \node[] at (0.5,1.5) {$v_u$};
    \node[] at (0.5,4.5) {$v_r$};
    
\end{tikzpicture}
\caption{}
\label{fig:dsp_consept:a}

\end{subfigure}
\begin{subfigure}{0.45\linewidth}
\begin{tikzpicture}[scale=0.6]
    
    \definecolor{high}{rgb}{1,0.3,0.3}
    
    \fill[black] (2,9) rectangle (3,8);
    \fill[black] (3,2) rectangle (4,0);
    \fill[black] (0,5) rectangle (6,4);
    
    \fill[high] (1,9) rectangle (2,7);
    \fill[high] (3,9) rectangle (4,7);
    \fill[high] (1,8) rectangle (4,7);
    
    \fill[high] (0,6) rectangle (6,5);
    \fill[high] (6,6) rectangle (7,3);
    \fill[high] (0,4) rectangle (7,3);
    
    \fill[high] (2,2) rectangle (3,0);
    \fill[high] (4,3) rectangle (5,0);
    \fill[high] (2,3) rectangle (4,2);
    
    \fill [pattern = north west lines, pattern color=blue] (0,2) rectangle (3,0);
    
    \draw (0,0) grid (8,9);

    \draw[thick, directed] (7.5, 4.5) -- (7.5,3.5);
    \draw[thick, directed] (7.5, 3.5) -- (6.5,2.5);
    \draw[thick, directed] (6.5, 2.5) -- (2.5,2.5);
    \draw[thick, directed] (2.5, 2.5) -- (1.5,1.5);
   
    \fill[green] (7.5,4.5) circle (0.3);
    \fill[red] (1.5,1.5) circle (0.3);
    \node[] at (7.5,4.5) {$\mathbf{A}$};
    \node[] at (1.5,1.5) {$\mathbf{B}$};
    
    \node[white] at (0.5,4.5) {$C_o$};
    \node[] at (0.5,2.5) {$C_f$};
    \node[] at (0.5,1.5) {$C_u$};
    \node[] at (0.5,3.5) {$C_r$};
    \node[white] at (2.8,0.5) {$C_{r+u}$};
    
\end{tikzpicture}
\caption{}
\label{fig:dsp_consept:b}
\end{subfigure}
\caption{D$^*_+$ outcome in planning a path from $\mathbf{A}$ to $\mathbf{B}$ in a partially known area where the unknown part is denoted as blue striped. As the robot traverse in the initially planned path, the robot discovers and register more known area and as such the overall path to be executed is online updated.}
\label{fig:dsp_consept}
\end{figure}

While the robotic platform is traversing, the mapping software is continuously updating the known map or $\mathbb{M}$, while at the same time it is expanding it to include the newly discovered areas and adding the previously unknown sections, while adjusting the incorrectly registered voxels as indicated in Figures~\ref{fig:dsp_consept:a} and~\ref{fig:dsp_consept:b}.

To accommodate for these changes in $\mathbb{M}$, the $\mathbb{G}$ is created dynamically so that it is capable of expanding and adjusting according to $\mathbb{M}$. In case D$^*_+$ is deployed in an area where dynamic obstacles exist, the $\mathbb{M}$ and subsequently the $\mathbb{G}$ will be updated and as such, $P$ will be updated accordingly, thus attempting to keep it safe and collision-free. 
The dynamic expansion of $\mathbb{G}$ in D$^*_+$ is to guarantee that $P$ will be updated in time to avoid a collision if obstacles are moving or the environment changes. Nevertheless, D$^*_+$ cannot replace a low-level local reactive obstacle avoidance module, in situations where the expansion time is longer.
Even if it would be fast enough is it still recommended to have a reactive safety component in the autonomy stack to ensure safety in the overall robotic mission. 

\begin{algorithm}
\caption{Decide if $\mathbb{G}$ will be updated or recreated}
\label{algo:1}
\SetAlgoLined
\textbf{Input:} New map $^nM$\\
Old map $^lM$\\
\textbf{Output:} grid $\mathbb{G}$\\

\eIf {size of $^nM$ $==$ size of $^lM$}{
    Algorithm~\ref{algo:update} ($^nM$) \Comment{If the same size update existing $\mathbb{G}$}
} {
    Algorithm~\ref{algo:create} ($^nM$) \Comment{Else create a new $\mathbb{G}$}
}
\end{algorithm}
\begin{algorithm}
\caption{Create grid $\mathbb{G}$}
\label{algo:create}
\textbf{Input:} Map $M$\\
\textbf{Output:} Grid $\mathbb{G}$

$g$ $\gets$ new array[size of $M$ : $C_u$]\\
\For {\textbf{each} $v$ \textbf{in} $M$}{
    \eIf {$v == v_o$}{ \Comment{If occupied set occupied and add traversal cost to $v$ in the proximity}\\
        $g[v] \gets C_o$\\
        Algorithm~\ref{algo:prox} ($v$)
    }{
        $g[v] \gets g[v] - C_u$ \\
        \Comment{If free remove cost for $C_u$ but keep any $C_r$}
    }
}

OctoGrid $\gets$ from $g$ \\
\Comment{Connect and search the OctoGrid}\\
$\mathbb{G}$ $\gets$ from OctoGrid \\


\end{algorithm}
\begin{algorithm}
\caption{Update occupancy grid $\mathbb{G}$}
\label{algo:update}
\textbf{Input:} New map $^nM$\\
Old map $^lM$\\
\textbf{Output:} Updated occupancy grid $\mathbb{G}$\\

\For {\textbf{each} $^nv$ \textbf{in} $^nM$}{
    \uIf {$C_{^nv} == C_o \textbf{ and } C_{^lv} \ne C_o$ }{
        $\mathbb{G}[^nv] \gets C_{^nv}$ \\
        Algorithm~\ref{algo:prox} ($^nv$) \\
        \Comment{Set proximity cost for surrounding voxels}
    } \uElseIf{$C_{^nv} == C_f \textbf{ and } C_{^lv} \geq C_u$}{
        \eIf{$ C_{^lv} == C_o$}{
            $\mathbb{G}[^nv] \gets C_f$ \\
            Algorithm~\ref{algo:fprox} ($^nv$) \Comment{Adding $C_r$ if needed}
        }{
            $\mathbb{G}[^nv] \gets C_{^lv} - C_u$ \\
            \Comment{Removing $C_u$ without changing $C_r$}
        }
    }
}
\end{algorithm}
\begin{algorithm}
\caption{Calculate $C_r$ for voxels within $r$ from $v_o$}
\label{algo:prox}
\textbf{Input:} $v$\\
\textbf{Output:} Updates in $\mathbb{G}$\\
\For {\textbf{each} $i$ \textbf{in} $[-r, r]$}{
\For {\textbf{each} $j$ \textbf{in} $[-r, r]$}{
\For {\textbf{each} $k$ \textbf{in} $[-r, r]$}{
    $v_r \gets \mathbb{G}[v_x+i, v_y+j, v_z+k$]\\
    $C_r \gets C_u / (i^2 + j^2 + k^2 + 1)$\\
    \uIf{$C_{vr} == C_o \textbf{ and } C_r > C_v$}{
        $\mathbb{G}[v] \gets C_r$
    }\uElseIf{$C_{vr} == C_o \textbf{ and } C_v \geq C_u \textbf{ and } C_r + C_u > C_v$}{
        $\mathbb{G}[v] \gets C_r + C_u$
    }
}
}
}

\end{algorithm}
\begin{algorithm}
\caption{Calculate $C_r$ when a $v$ is discovered to be free}
\label{algo:fprox}
\textbf{Input:} $v$\\
\textbf{Output:} Updates in $\mathbb{G}$\\
$C_r \gets 0$\\
\For {\textbf{each} $i$ \textbf{in} $[-r, r]$}{
\For {\textbf{each} $j$ \textbf{in} $[-r, r]$}{
\For {\textbf{each} $k$ \textbf{in} $[-r, r]$}{
    $v_r \gets \mathbb{G}[v_x+i, v_y+j, v_z+k]$\\
    \If{$C_{vr} == C_o$}{
        $C \gets c_u / (i^2 + j^2 + k^2 + 1)$\\
        \If{$C > C_r$}{
            $C_r \gets C$
        }
    }
}
}
}
$\mathbb{G}[v] \gets C_r$ 
\end{algorithm}

\section{Experiments}\label{sec:exp}
Three sets of experiments were executed: a) with the Boston Dynamics Spot robot, used for 2D path planning, b) with a UAV for 3D path planning c) comparison of 3D planning using the UAV.
The duration of the experiments was varying among the performed experiments between approximately \unit[45]{s} and \unit[680]{s}, where the robots had an average moving velocity of approximately \unit[0.7]{m/s}.
To execute the experiments that evaluated the D$^*_+$'s capability to plan for a safe path that a robot can traverse, multiple additional components are needed for enabling an overall autonomous robotic mission that indirectly affects the overall performance mission. As such, in the rest of this Section, we will provide a comprehensive overview of the robotic configurations and the software components that were involved in the experimental evaluations.

\subsection{2D path planning: The Boston Dynamics Spot use case}

In this work, the 2D path planning capabilities of D$^*_+$ are tested with the BD Spot robot, which is equipped with a 3D lidar, an Inertial Measurement Unit (IMU), and an onboard computer. The software architecture used is depicted in Figure~\ref{fig:spot_artc}. In this software configuration, the Google Cartographer~\cite{45466} was utilized is a Simultaneous Localisation and Mapping (SLAM) algorithm being able to create a 2D gird map $\mathbb{M}_2$ and the corresponding state estimation information. The information from the state estimation uses the current position of the robot, as the starting position $\mathbf{A}$ that together with $\mathbb{M}_2$, and the goal position $\mathbf{B}$ are used by D$^*_+$ to plan a path $P$. Moreover, a position controller takes $P$ and gives corresponding velocity commands to Spot's built-in low-level motion controller that makes Spot move. In the experiments with Spot, a prebuild $\mathcal{M}_2$ of the area was provided and thus allowing for a longer, than the sensor range, $P$ to be planned from the beginning of the mission.
\input{tizk/spot_artchetecture}
A more complete and exhaustive presentation of utilized Spot's autonomous capability is presented in~\cite{koval2022experimental}.

\begin{figure}[htbp!]
\centering
\resizebox{0.9\textwidth}{!}{%
\begin{tikzpicture}
    \node at (0,0) {\includegraphics[width=\textwidth]{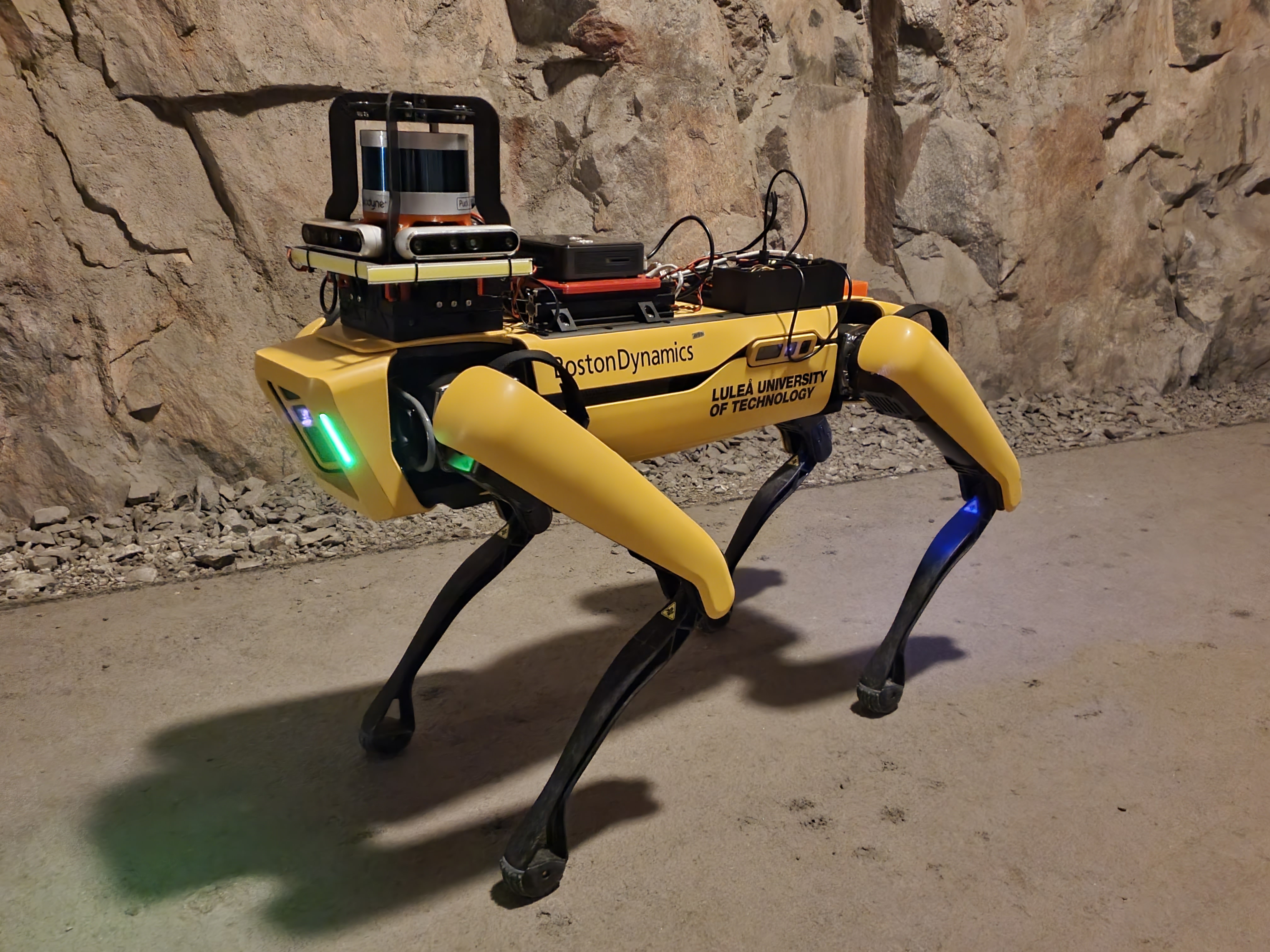}};
    
     \draw[-stealth, white] (-5.0, 2.9) -> (-2.0,2.9);
     \draw[-stealth, white] (-5.0, 1.7) -> (-2.0,1.7);
     \draw[-stealth, white] (-0.4, 2.7) -> (-0.4,2.3);
     \draw[-stealth, white] (1.3, 2.8) -> (1.3,2.2);
     
     \node[white] at (-4.3, 3.5) {Velodyne};
     \node[white] at (-4.3, 3.2) {VLP16};
     \node[white] at (-4.3, 2.3) {Vectornav};
     \node[white] at (-4.3, 2.0) {VN-100};
     \node[white] at (-0.4, 3.0) {Intel NUC};
     \node[align=center,white] at (1.5, 3.3) {Auxiliary};
     \node[align=center,white] at (1.5, 3.0) {connections};
\end{tikzpicture}
}
    \caption{Spot equipped with the attached sensors for the autonomy navigation and computational power. In this view perspective, the Vectornav IMU is placed under the Velodyne lidar and is not visible in the Figure.}
    \label{fig:Spot-autonomy-package}
\end{figure}

\subsection{3D path planning: the UAV use case}
\begin{figure}
    \centering
    \resizebox{0.9\textwidth}{!}{%
\begin{tikzpicture}
    \node at (0,0) {\includegraphics[width=\textwidth]{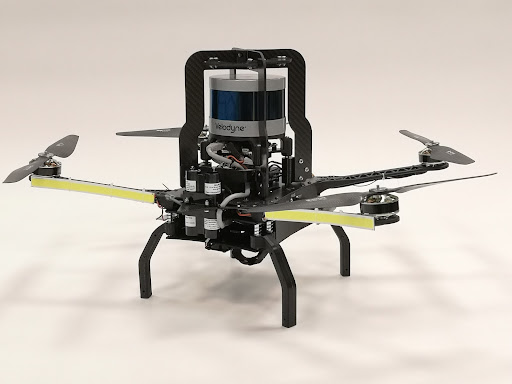}};
    
     \draw[-stealth] (-3.5, 2.0) -> (-1.1,2.0);
     \draw[-stealth] (2.5, 0.8) -> (0.4,0.8);
     \draw[-stealth] (-5.1, -0.5) -> (-1.4,-0.5);
     \draw[-stealth] (4.0, -1.0) -> (1.0,-1.0);

     \node[] at (-2.8, 2.5) {Velodyne};
     \node[] at (-2.8, 2.2) {VLP16};
     \node[] at (-3.8, -0.8) {Single beam lidar};
     \node[] at (2.0, 1.0) {IMU};
     \node[] at (3.2, -1.2) {Intel NUC};
\end{tikzpicture}
}
    \caption{The UAV used for the 3D path planning experiments.}
    \label{fig:shafter}
\end{figure}

For the 3D path planning experiments, a custom-built UAV was used as depicted in Figure~\ref{fig:shafter}. In this case, the D$^*_+$ architecture operates in the 3-dimensional space and requires as input the current position $\mathbf{A}$, a goal position $\mathbf{B}$ and an occupancy map $\mathbb{M}_3$, while it provides the path $P_{\mathbf{A} \to \mathbf{B}}$ as an output. The autonomous navigation of the robot is based on the path generated from D$^*_+$, which essentially coordinates the motion of the robot, while other modules, such as the state estimation, and a Non-Linear Model Predictive Controller (NMPC)~\cite{9625659} that uses an artificial potential field algorithm~\cite{lindqvist2021compra} to reactively avoid obstacles, are part of the overall mission execution. The UAV's primary sensor in the autonomy stack is a Velodyne Puck Lite lidar, which is used alongside an IMU to provide state estimation information on the current position $\mathbf{A}$. In this work, the $\mathbf{A}$ is calculated based on LIO-SAM~\cite{liosam2020shan}, while the 3D occupancy map $\mathbb{M}_3$ is generated based on the Octomap~\cite{hornung13auro} framework.


In these experiments, the artificial potential fields are only used as an extra safety layer to avoid collisions when flying close to obstacles (e.g. less than 50cm relative distance to an obstacle), where the designed behavior is to prioritize the influence of the potential fields to directly affect the movement since the path, in that case, is considered to be unsafe. Furthermore, the software architecture for the UAV is visualized in Figure~\ref{fig:shafter_arch}.
\input{tizk/shafter_artchetecture}
In all the 3D tests, no information was given a-priori, meaning that the UAV explored $\mathcal{M}$ during the mission, thus utilizing both related map updates and expansions. During the experiments, the waypoints were manually set as the mission was evolving and the desired waypoints were reached, while after some time a waypoint was commanded to set a return to the starting location.

\subsection{Graph-based Exploration Planner 2.0}
The Graph-based Exploration Planner 2.0 was selected to do a comparison with the proposed method because of its proven capabilities in the DARPA subterranean challenge and the possibility to run it in a waypoint navigation mode, which is similar to how the D$^*_+$ is used. In general, there are considerable differences between Gbplanner and D$^*_+$, which may bias the evaluation results, depending on the experimental setup.
In order to achieve a representative comparison in Gbplanner, the collision model was adjusted to match the UAV size.

\section{Experimental Results}\label{sec:result}
\subsection{2D Path planning}

\begin{figure}[htbp!]
    \centering
    \resizebox{0.9\textwidth}{!}{%
    \begin{tikzpicture}[rotate=90]
        \node at(0,0) {\includegraphics[width=\textwidth, angle=90]{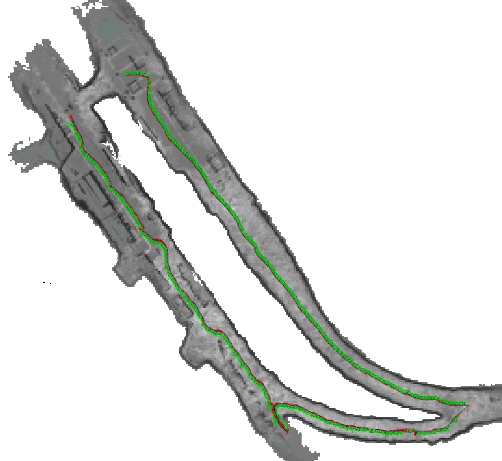}};
        
        \fill[green] (-3.0,3.75) circle (0.1);
        \fill[blue] (5.3,-4.2) circle (0.1);
        \fill[blue] (0.9,-4.9) circle (0.1);
        
        \fill[blue] (-1.2,-2.2) circle (0.1);
        \fill[blue] (-3.35,1.0) circle (0.1);
        
        \fill[blue] (-2.6,0.0) circle (0.1);
        \fill[red] (-4.3,2.6) circle (0.1);
        
        \node[white] at (-3.3, 3.95) {$\mathbf{A}$};
        \node[white] at (5.65, -4.2) {$\mathbf{B}_{1}$};
        \node[white] at (-1.2, -2.7) {$\mathbf{B}_{2,4}$};
        \node[white] at (0.9, -5.3) {$\mathbf{B}_{3}$};
        \node[white] at (-2.3, 0.1) {$\mathbf{B}_{5}$};
        \node[white] at (-3.5, 0.7) {$\mathbf{B}_{6}$};
        \node[white] at (-4.0, 2.7) {$\mathbf{B}_{7}$};
        
    \end{tikzpicture}
    }
    \caption{Sequential $\mathbf{B}$ were set in a large subterranean tunnel. Spot navigates to each $\mathbf{B}$ and traverses about $\unit[335]{m}$ in total. In this experiment, where the map is known a priori, D$^*_+$ shows an increased capability to plan longer paths and excellent replanning capabilities. In this case, $\mathbf{B}_3$ was set before Spot reached $\mathbf{B}_2$, thus forcing a replanning. When $\mathbf{B}_3$ was reached where a new point $\mathbf{B}_4$ was set at approximately the same point as $\mathbf{B}_2$.}
    \label{fig:tunnel-areas-lkab}
\end{figure}

Among the platforms that were used in our experiments, Spot robot has a significantly longer battery life than the UAV and can explore larger subterranean areas. Thus, in this article, the Spot robot was used to evaluate the D$^*_+$ 2D performance. This allowed us to evaluate path planning in larger maps and over longer distances in comparison with the UAV. Worth noting that the computational load for 2D path planning was significantly lower than for 3D.
In this case, the experiments with Spot showed that D$^*_+$ is able to plan long paths that were approximately \unit[80]{m} at a time and without collisions. One such experiment, where a multiple waypoint mission was executed, is presented in Figure~\ref{fig:tunnel-areas-lkab}, where a total of $7$ waypoints, were reached progressively. in this case, Spot was capable of navigating to each of the waypoints without collisions.

\begin{figure}
    \centering
    \includegraphics[width=0.9\textwidth]{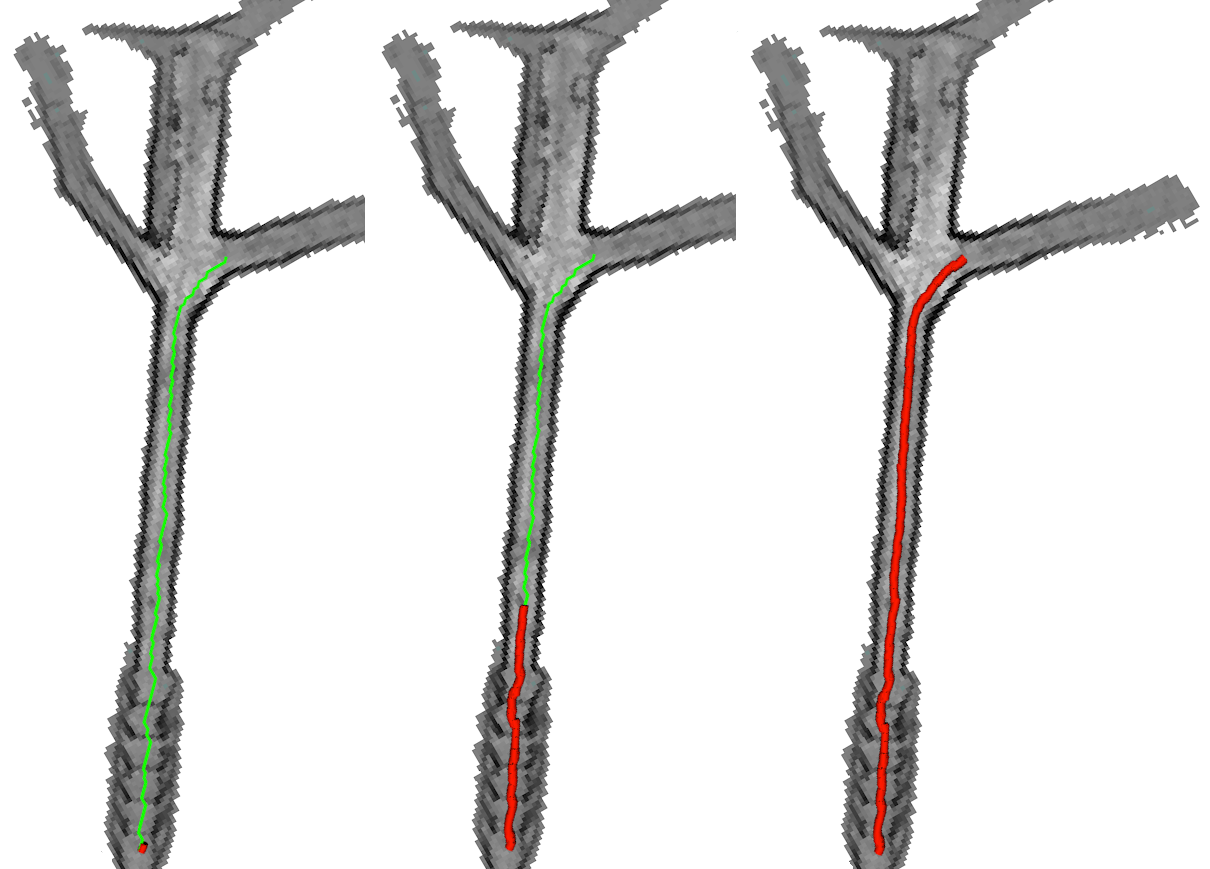}
    \caption{Spot navigates along the planned path shown in green, and the traversed path shown in red.}
    \label{fig:spotShafter}
\end{figure}

Another experiment was focusing on the use case where Spot had to traverse approximately $\unit[80]{m}$ in a more narrow tunnel. Figure~\ref{fig:spotShafter} shows how Spot can follow the path planned by D$^*_+$ as it has suitable distances to both walls.

\begin{figure}
    \centering
    \includegraphics[width=0.9\textwidth]{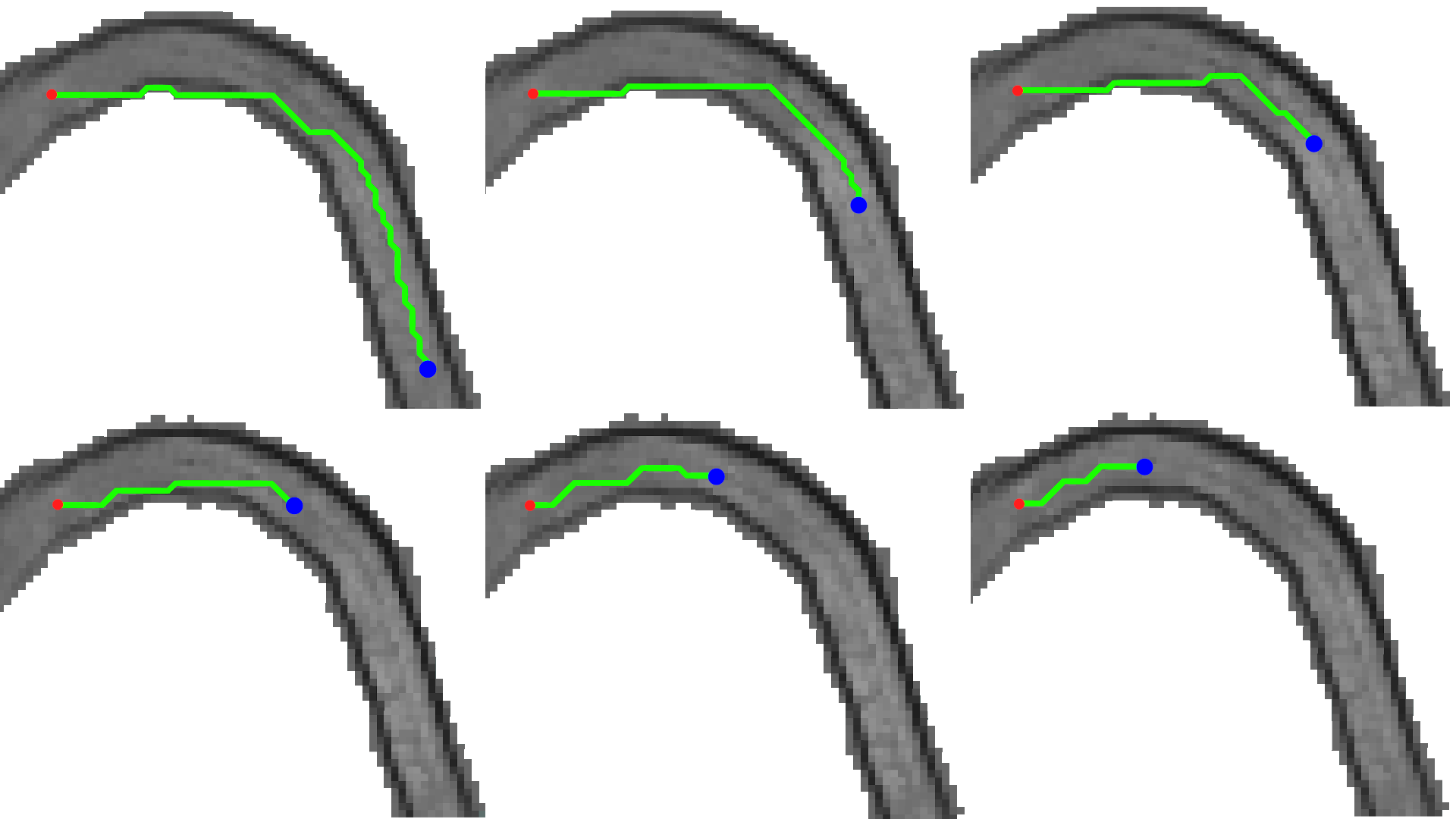}
    \caption{As Spot traverse the D$^*_+$ planned path from the blue dot to the red dot, more information on the surrounding environment is provided. Thus the planner updates the current $\mathbb{M}$ causing D$^*_+$ to reactively re-plan and adjust the overall path.}
    \label{fig:update}
\end{figure}

When $\mathbb{M}$ updates it will generate new conditions for D$^*_+$ to plan paths, therefore D$^*_+$ will constantly perform an update of the path to keep it safe. In Figure~\ref{fig:update} it is presented an example of how the path may be re-planed as new information about the environment is acquired. As such, the starting path is planned initially inside a wall but as it is discovered in the corresponding map updates, the path is interfering with a wall and thus reactively D$^*_+$ performs a re-planning of the path to stay away from it and in a safe operation.

\begin{figure}
    \centering
    \includegraphics[width=0.7\textwidth]{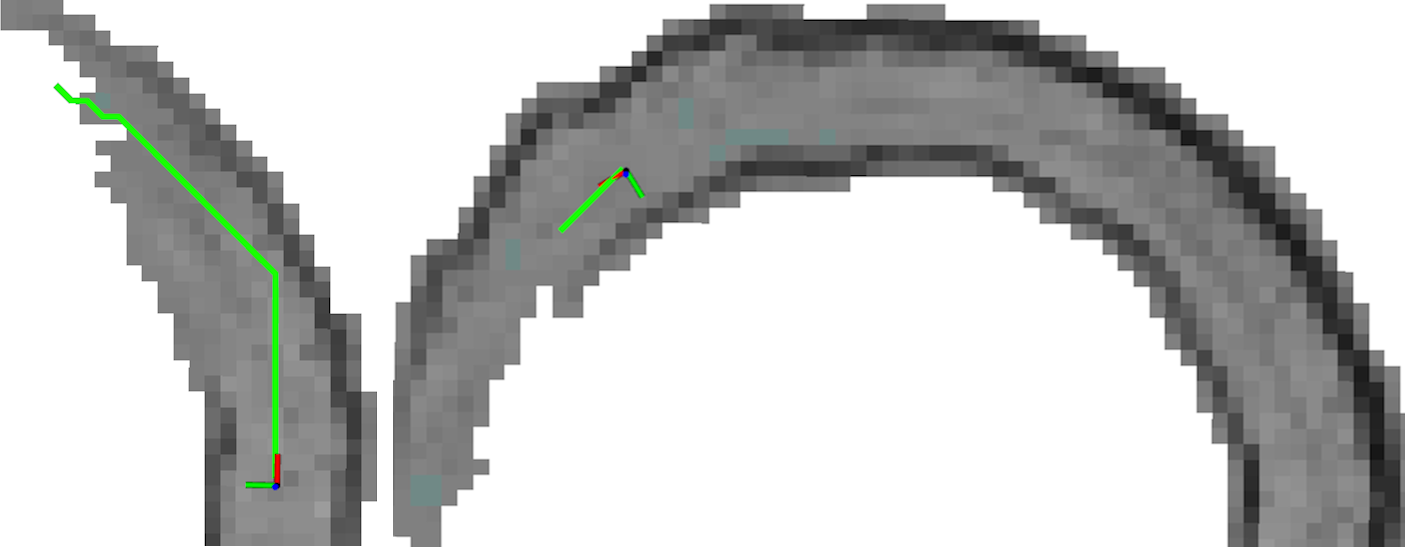}
    \caption{On the left path planned until the edge of the known map and on the right new path planned towards the goal after a map expansion of the same area. In both cases, the path is denoted by the solid green line.}
    \label{fig:expand}
\end{figure}

Another merit of the proposed D$^*_+$ is the sequential map and graph update capability, which allows planning or re-plan paths towards $\mathbb{B}$ once the robot discovers more parts of the area that navigates along. Figure~\ref{fig:expand} depicts the map expansion capability with Spot traversing in the tunnel, wherein the first case the path is planned until the edge of the known map and then a new path is planned in the expanded map of the same area to reach another goal during the mission.

\subsection{3D Path Planning}
\begin{figure}
\centering
\begin{subfigure}{.32\textwidth}
  \centering
  \includegraphics[width=0.9\linewidth]{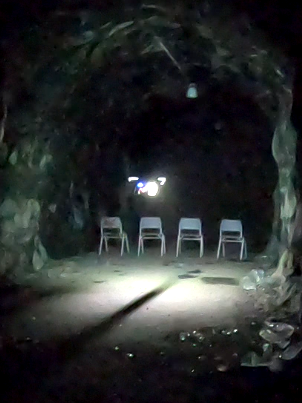}
  \caption{Low obstacle.}
  \label{fig:low}
\end{subfigure}%
\begin{subfigure}{.32\textwidth}
  \centering
  \includegraphics[width=0.9\linewidth]{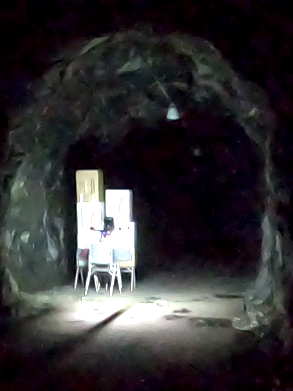}
  \caption{Obstacle to the side.}
  \label{fig:side}
\end{subfigure}
\begin{subfigure}{.32\textwidth}
  \centering
  \includegraphics[width=0.9\linewidth]{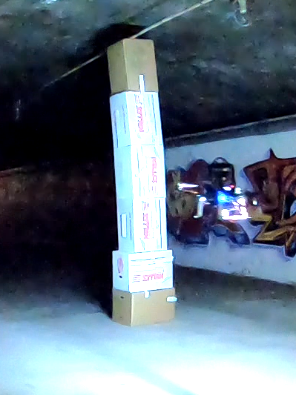}
  \caption{Pillar in the void.}
  \label{fig:void}
\end{subfigure}
\caption{Snapshots from three of experiments executed with the UAV.}
\label{fig:test}
\end{figure}

D$^*_+$ 3D planning capabilities were also tested in three experiments in a cave tunnel where it was challenged to plan $P$ along a) a low blockage as depicted in Figure~\ref{fig:low}, b) a tall obstacle to the side making a narrow passage as in Figure~\ref{fig:side}, and c) around a pillar as depicted in Figure~\ref{fig:void}.

\begin{figure}
    \centering
    \includegraphics[width=0.9\textwidth]{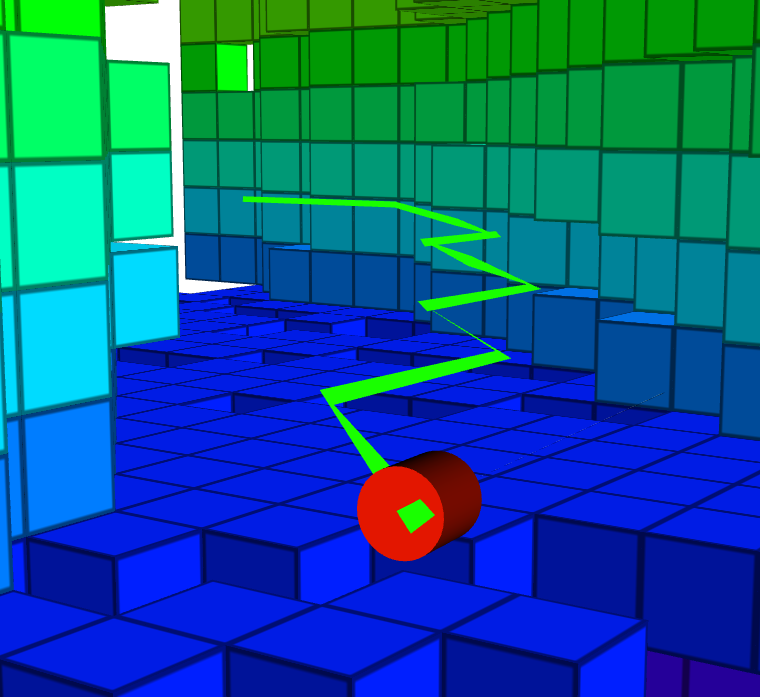}
    \caption{D$^*_+$ plans a path around a bend while keeping a suitable distance to the inside wall.}
    \label{fig:3d_long}
\end{figure}

In the tunnel corner experiment, the D$^*_+$ planned $P$ to follow the bend with a suitable safe marginal to the walls. As shown in Figure~\ref{fig:3d_long} D$^*_+$ is planning a safe path following the tunnel where only the center point of voxels is selected as waypoints in the path, resulting in a zig-zag path.

\begin{figure}
    \centering
    \includegraphics[width=0.9\textwidth]{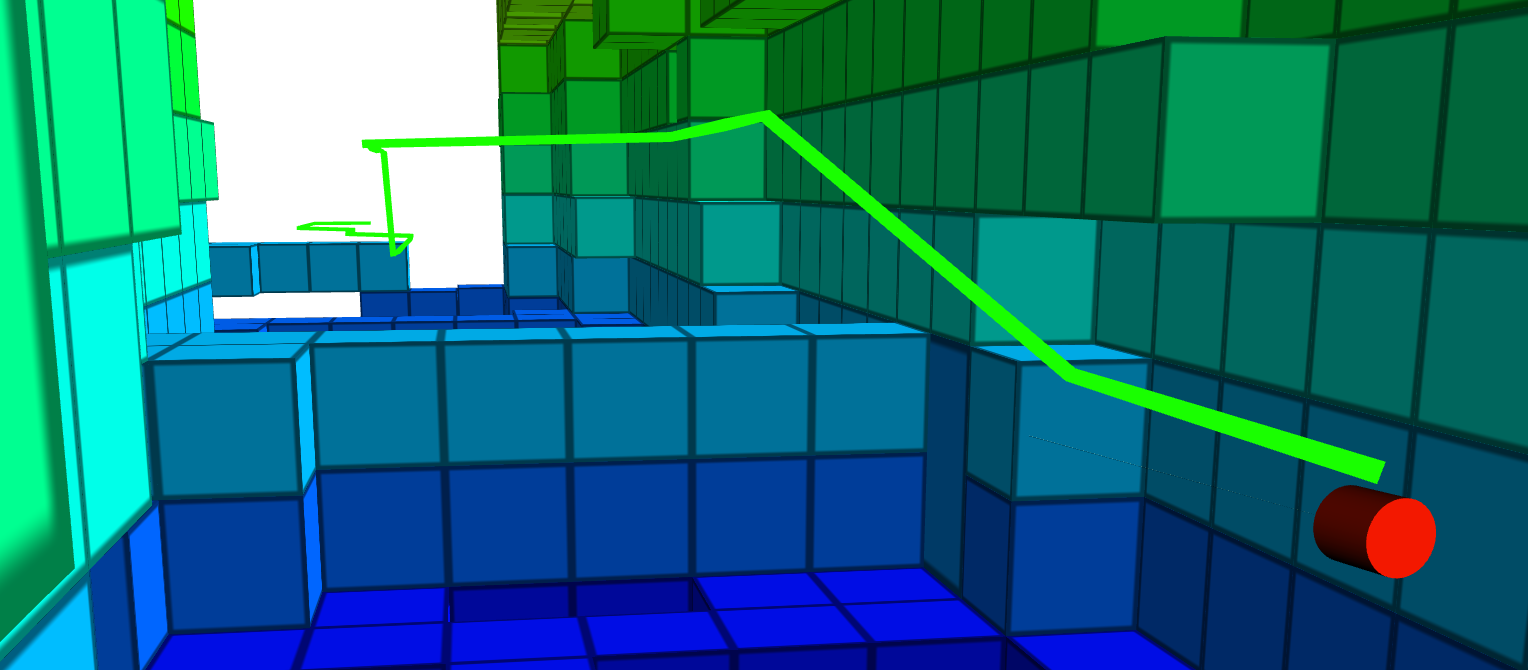}
    \caption{D$^*_+$ path planning with a low blockage so that the UAV needs to fly higher to keep a suitable safety marginal to the blockage.}
    \label{fig:over}
\end{figure}

D$^*_+$ was also challenged with a low-height obstacle that could potentially risk a collision with the UAV's landing gear if it keeps its current altitude. In this case, $P$ is planned with an altitude change in order to keep the path safe, as depicted in Figure~\ref{fig:over}.

\begin{figure}
    \centering
    \includegraphics[width=0.9\textwidth]{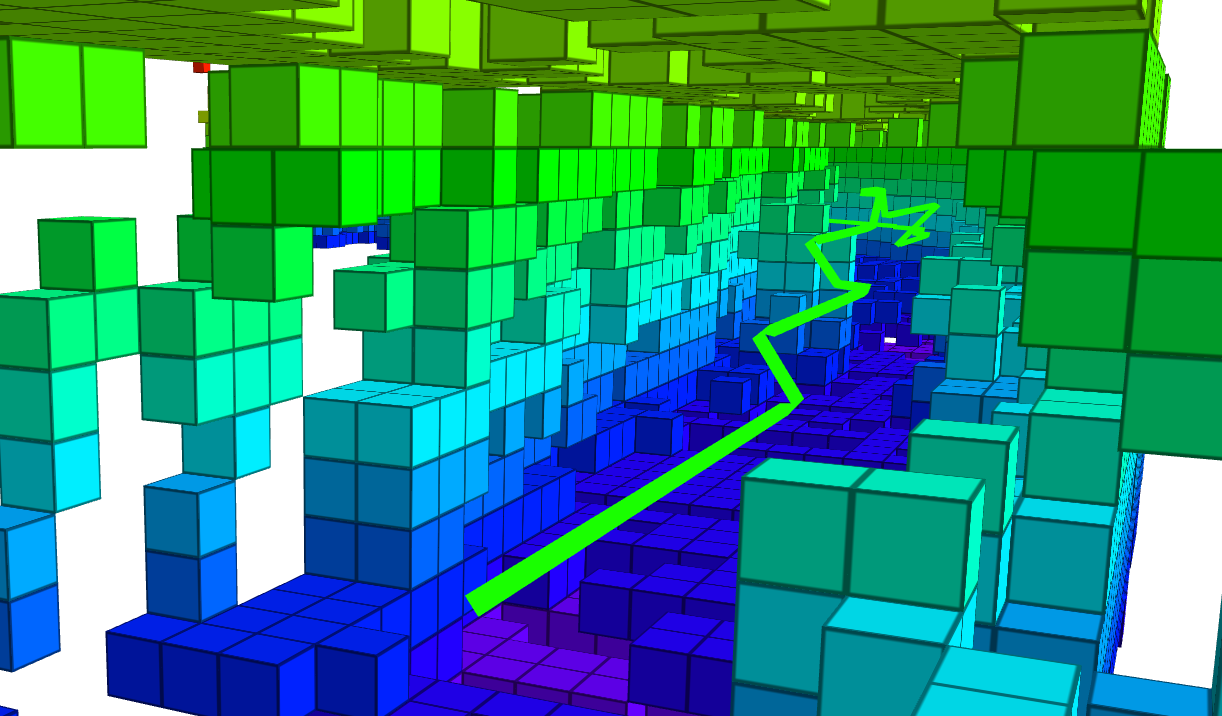}
    \caption{D$^*_+$ is capable of planning a path from beyond the line of sight to the end of the green line while having the desired safety margin throughout the whole path.}
    \label{fig:distance}
\end{figure}

At this point, it should be also mentioned that when D$^*_+$ is tasked to plan a path to a point beyond the line of sight, the algorithm is still capable of providing a path that is inherently safe and short as possible. This performance is depicted in Figure~\ref{fig:distance} whereas it is shown, the free space gap between the obstacle to the left and the right wall, was very narrow for the UAV to pass through them, and thus a higher path has been selected.

\begin{figure}
    \centering
    \includegraphics[width=0.9\textwidth]{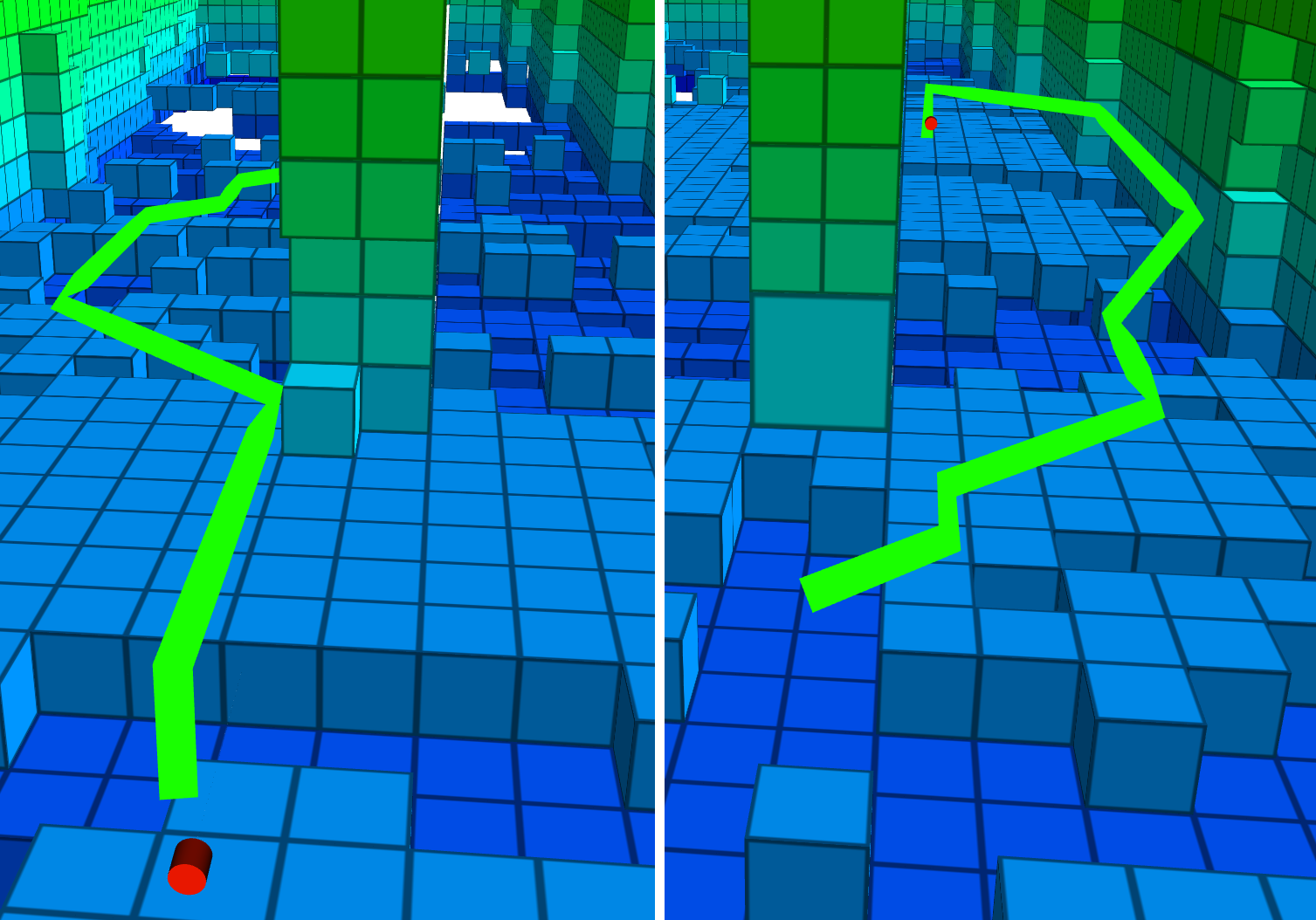}
    \caption{D$^*_+$ is tasked to plan a path from the red dot to the end of the green line. The resulting path keeps proper safety margins that are relative to the pillar so that the UAV can traverse it safely.}
    \label{fig:pilar}
\end{figure}

In the last experimental test, the D$^*_+$ is tasked to plan a path from one side of a pillar to the other and then back again.
This experiment shows that D$^*_+$ takes the shortest path outside the risky area from the pillar, as also depicted in Figure~\ref{fig:pilar}.

\begin{figure}
    \centering
    \includegraphics[width=0.9\textwidth]{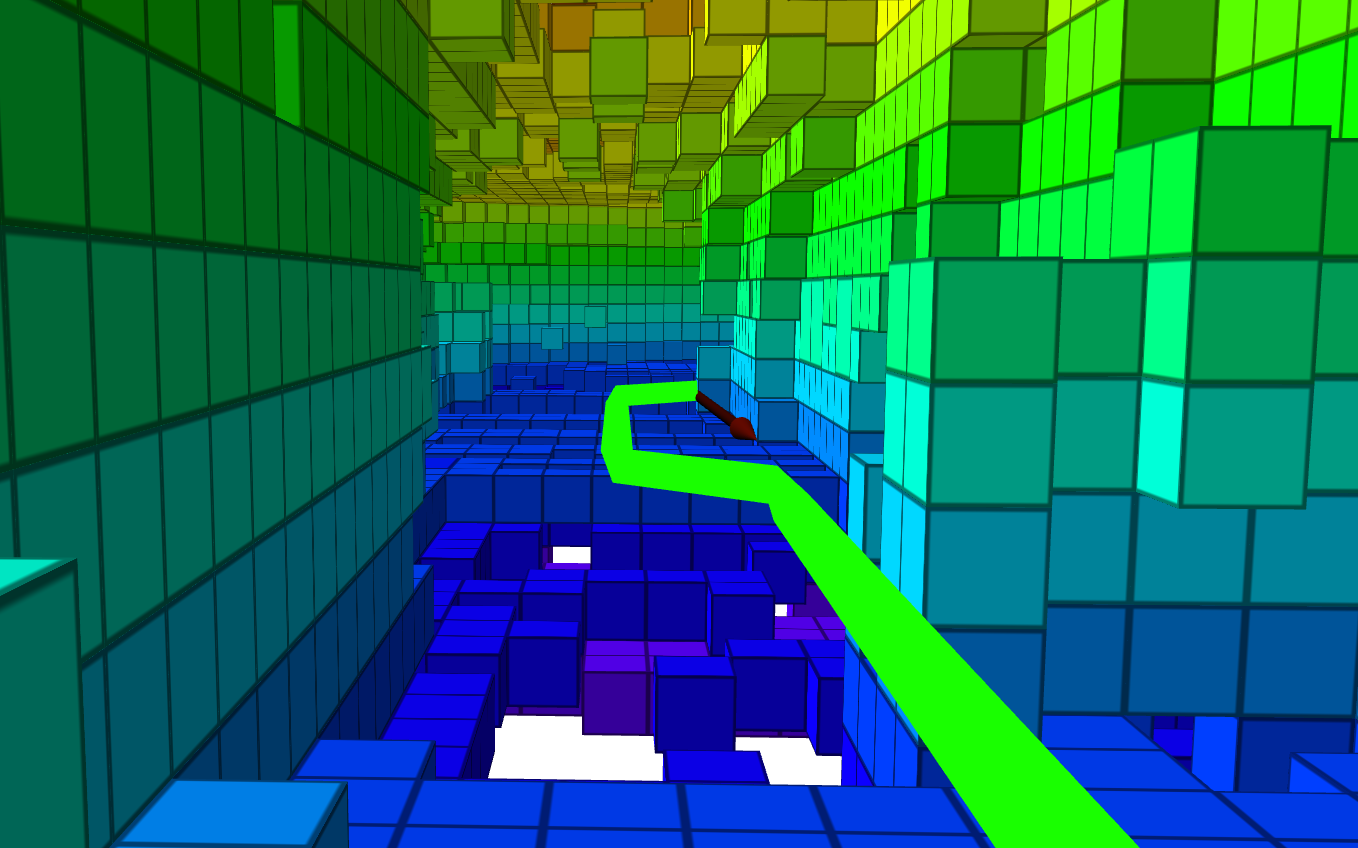}
    \caption{A UAV avoids colliding with an obstacle on the side by planning $P$ (the green line) in the center of the opening.}
    \label{fig:3d_side}
\end{figure}

During the experiment, with an obstacle at the side of the tunnel, D$^*_+$ planned $P$ that was in the center of the opening as shown in Figure~\ref{fig:3d_side}.
When passing the obstacle, the remaining planned path converges again to the center of the opening, while keeping a safe distance to all sides.

\begin{figure}
    \centering
    \includegraphics[width=0.9\textwidth]{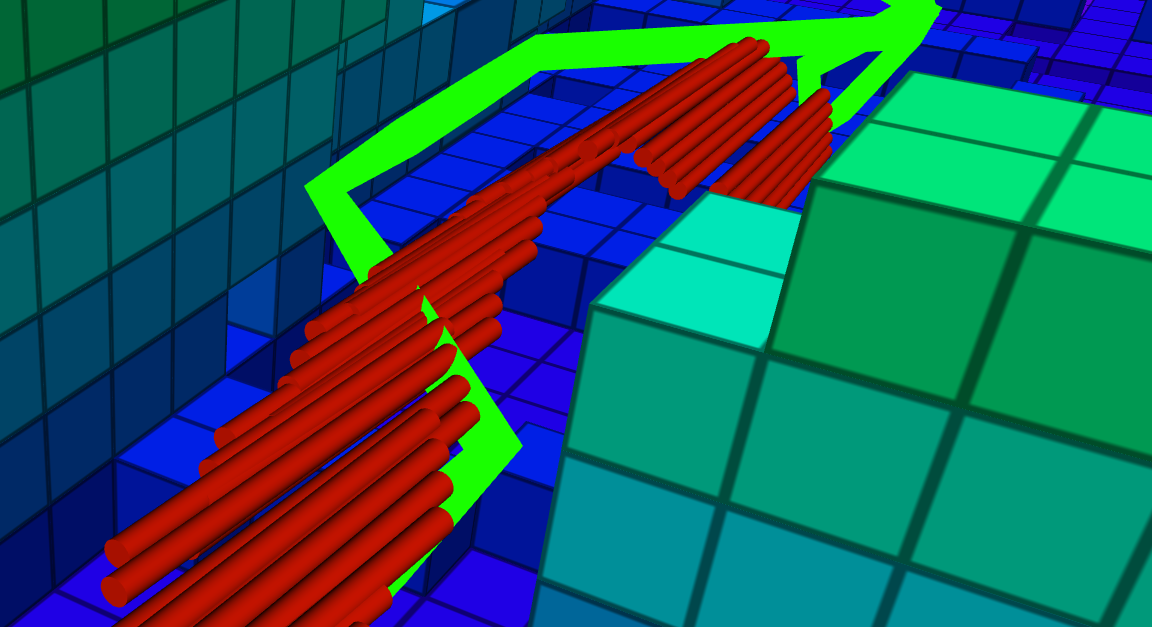}
    \caption{An instance from the autonomous mission and corresponding path planning when the UAV took a too-tight turn around an obstacle and caused the artificial potential field to avoid a collision. The green line is $P$ and the red bars are the path followed by the UAV.}
    \label{fig:3d_side_cuting}
\end{figure}

During all the experimental trials with the UAV, there have been only four times that the potential fields have been triggered to avoid a collision, while two of them were barely noticeable, only just clipping a corner of the protected area, while it should be noted that: a) the safety would have been guaranteed even without the potential fields, b) one triggering  was because $\mathbf{B}$ were set to close to a wall, and c) the fourth time was during the return on the side obstacle, where $P$ were planned tightly around that. In the final case, the path following had a slightly too long look ahead distance, thus causing it to shortcut the corner in a more intense approach, as it can be seen in Figure~\ref{fig:3d_side_cuting}.
The latest was a dangerous passage close to the obstacle but it was not due to a fault in the D$^*_+$ planner, rather than a weakness in the path following the autonomy stack. In this case, the easiest solution would have been to increase $r$ so $P$ are planned with the proper safe margins for these types of corner-cutting as well.


\subsection{Comparison with Graph-based Exploration Planner 2.0}
The experiments used for path planners comparison were carried out in the curved subterranean tunnel in Lule\r{a}, Sweden. To evaluate their performance the following metrics were used: CPU and memory usage, as well as an assessment of the planned path.

\begin{figure}
\begin{subfigure}{.49\textwidth}
  \centering
  \includegraphics[width=0.9\linewidth]{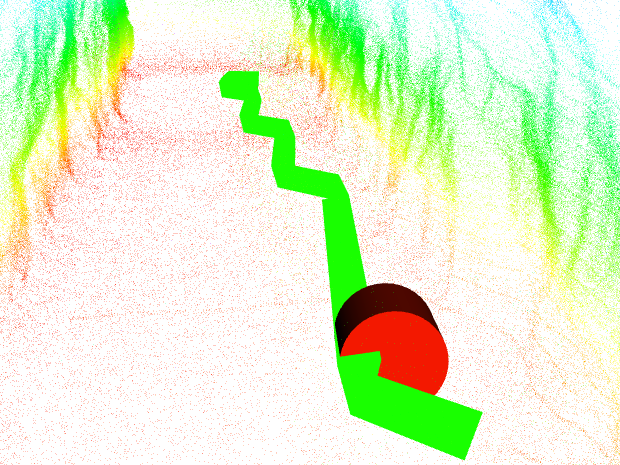}
  \caption{The path D$^*_+$ planned (green) when doing the comparison experiments. The path is as straight as voxel-based planning can be.}
  \label{fig:comparison:dsp}
\end{subfigure}%
\hfill
\begin{subfigure}{.49\textwidth}
  \centering
  \includegraphics[width=0.9\linewidth]{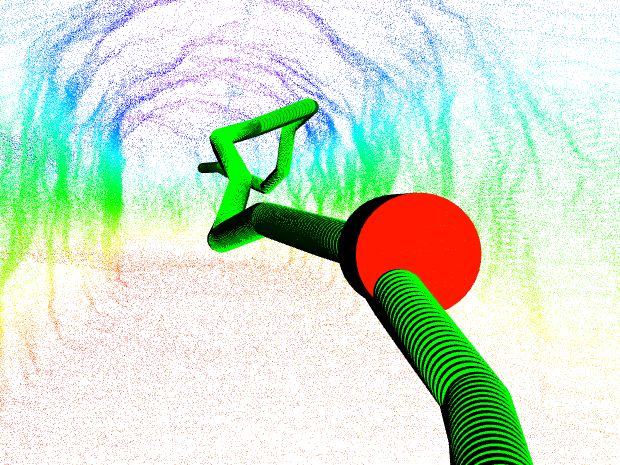}
  \caption{The path Gbplanner planned (green) when doing comparison experiments. Note how the path is curving up and around for no apparent reason.}
  \label{fig:comparison:gb}
\end{subfigure}%
    \caption{Differences in planned paths between D$^*_+$ and Gbplanner are visible when both planners plan in the same area under the same conditions.}
    \label{fig:comparison}
\end{figure}

Both planners have been successfully planning a path to traverse the tunnel but the path characteristics differ significantly (see Figure~\ref{fig:comparison}). D$^*_+$ is planning the shortest path from voxel to voxel all the way to the end voxel, resulting in a strait path with no unnecessary detours. Gbplanner on the other hand is an RRT-based planner, which chooses a path leading left and right, up and down to reach a target that is accessible through a straight line. This approach produces less safe paths, like for example when the path goes close to small objects or walls as it is shown in Figure~\ref{fig:comparison:gb}.
\\\\
Based on our field experience, small structures are not guaranteed to be seen by the mapping algorithm so as a safety feature for the UAV we were using the artificial potential field.
It is likely that Gbplanner would have succeeded in any way but the tendency to approach walls and small objects were increasing the risk of collisions. Moreover, these unnecessary movements will reduce UAV's battery life and shorten the maximum distance of the mission.

When we are considering the computational load, the proposed D$^*_+$ is mainly utilizing the CPU for building the graph, which is correlated with the memory usage increases as depicted in Figure~\ref{fig:mem_comp}.
During planning and the $\mathbb{G}$ updating, the CPU load is at approximately $\unit[10]{\%}$ of a single core. When it comes to resource efficiency, D$^*_+$ is prone to high memory consumption. For example, in the evaluation test, D$^*_+$ used $\unit[3.5-7.0]{GB}$ of RAM, while Gbplanner used only $\unit[0.2-0.3]{GB}$ of RAM, as shown in Figure~\ref{fig:mem_comp}.
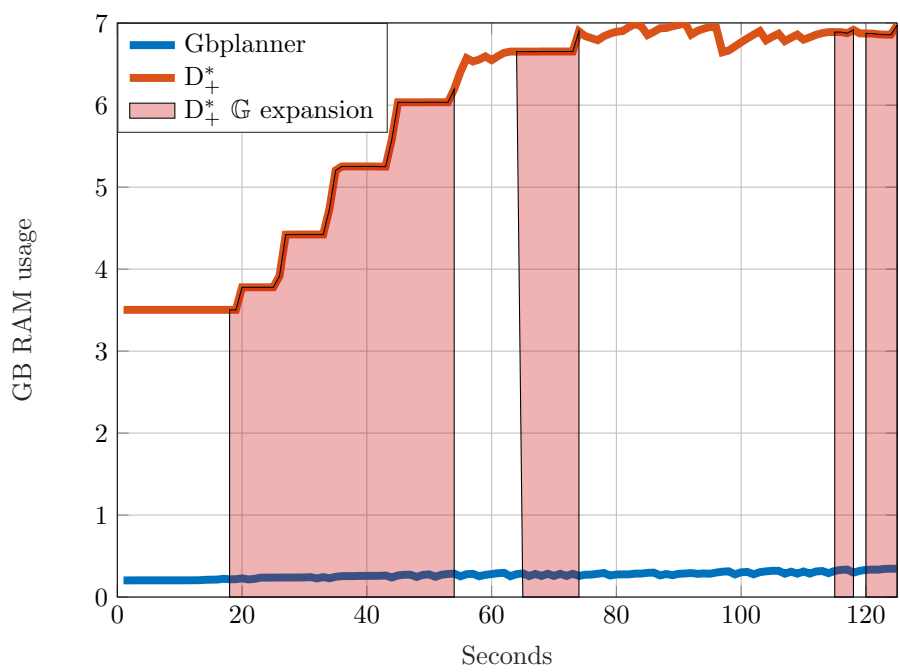
\begin{figure}
    \center
    \setlength\fwidth{0.9\columnwidth}
%
%
\definecolor{mycolor1}{rgb}{0.00000,0.44700,0.74100}%
\definecolor{mycolor2}{rgb}{0.85000,0.32500,0.09800}%
\begin{tikzpicture}

\begin{axis}[%
width=0.95\fwidth,
height=0.7\fwidth,
at={(0.9\fwidth,0.9\fwidth)},
scale only axis,
xmin=0,
xmax=125,
xlabel style={font=\color{white!15!black}},
xlabel={Seconds},
ymin=0,
ymax=7.003295744,
ylabel style={font=\color{white!15!black}},
ylabel={GB RAM usage},
axis background/.style={fill=white},
xmajorgrids,
ymajorgrids,
legend style={at={(0.00,1.0)}, anchor=north west, legend cell align=left, align=left, draw=white!15!black}
]
\addplot [color=mycolor1, line width=3.0pt]
  table[row sep=crcr]{%
1	0.202989568\\
2	0.202989568\\
3	0.203907072\\
4	0.203907072\\
5	0.203907072\\
6	0.203907072\\
7	0.203915264\\
8	0.203915264\\
9	0.203915264\\
10	0.203915264\\
11	0.203915264\\
12	0.204378112\\
13	0.20475904\\
14	0.209813504\\
15	0.212086784\\
16	0.212774912\\
17	0.222994432\\
18	0.216743936\\
19	0.21764096\\
20	0.228233216\\
21	0.214441984\\
22	0.221487104\\
23	0.236670976\\
24	0.236761088\\
25	0.237600768\\
26	0.237723648\\
27	0.237891584\\
28	0.23810048\\
29	0.238317568\\
30	0.239181824\\
31	0.242937856\\
32	0.226770944\\
33	0.243286016\\
34	0.229199872\\
35	0.246657024\\
36	0.255143936\\
37	0.2553856\\
38	0.255463424\\
39	0.25798656\\
40	0.257998848\\
41	0.258695168\\
42	0.258895872\\
43	0.262905856\\
44	0.23816192\\
45	0.263315456\\
46	0.271859712\\
47	0.272289792\\
48	0.244977664\\
49	0.270409728\\
50	0.276168704\\
51	0.246996992\\
52	0.272420864\\
53	0.281595904\\
54	0.285892608\\
55	0.25307136\\
56	0.278872064\\
57	0.282853376\\
58	0.253509632\\
59	0.270913536\\
60	0.279719936\\
61	0.290906112\\
62	0.295473152\\
63	0.253853696\\
64	0.279728128\\
65	0.288804864\\
66	0.255586304\\
67	0.28178432\\
68	0.257572864\\
69	0.2837504\\
70	0.257572864\\
71	0.283758592\\
72	0.257581056\\
73	0.283766784\\
74	0.257564672\\
75	0.272605184\\
76	0.274055168\\
77	0.283201536\\
78	0.293584896\\
79	0.262631424\\
80	0.275103744\\
81	0.276627456\\
82	0.276832256\\
83	0.285769728\\
84	0.285970432\\
85	0.295497728\\
86	0.297709568\\
87	0.26437632\\
88	0.281481216\\
89	0.266428416\\
90	0.283652096\\
91	0.2870272\\
92	0.293122048\\
93	0.282800128\\
94	0.28807168\\
95	0.284045312\\
96	0.298897408\\
97	0.309293056\\
98	0.313839616\\
99	0.27590656\\
100	0.302567424\\
101	0.305946624\\
102	0.277327872\\
103	0.30328832\\
104	0.31371264\\
105	0.32004096\\
106	0.320221184\\
107	0.284749824\\
108	0.308662272\\
109	0.285544448\\
110	0.311676928\\
111	0.28561408\\
112	0.311836672\\
113	0.317640704\\
114	0.29110272\\
115	0.318877696\\
116	0.331157504\\
117	0.334876672\\
118	0.29681664\\
119	0.319705088\\
120	0.331460608\\
121	0.334123008\\
122	0.334168064\\
123	0.34357248\\
124	0.345812992\\
125	0.345874432\\
};
\addlegendentry{Gbplanner}

\addplot [color=mycolor2, line width=3.0pt]
  table[row sep=crcr]{%
1	3.502964736\\
2	3.502964736\\
3	3.502964736\\
4	3.502964736\\
5	3.502964736\\
6	3.502964736\\
7	3.502964736\\
8	3.502964736\\
9	3.502964736\\
10	3.502964736\\
11	3.502964736\\
12	3.502964736\\
13	3.502964736\\
14	3.502964736\\
15	3.502964736\\
16	3.502964736\\
17	3.502964736\\
18	3.502964736\\
19	3.502964736\\
20	3.77843712\\
21	3.77843712\\
22	3.77843712\\
23	3.77843712\\
24	3.77843712\\
25	3.77843712\\
26	3.915767808\\
27	4.420214784\\
28	4.422647808\\
29	4.422647808\\
30	4.422647808\\
31	4.422647808\\
32	4.422647808\\
33	4.422647808\\
34	4.733210624\\
35	5.204533248\\
36	5.252653056\\
37	5.252653056\\
38	5.252579328\\
39	5.252542464\\
40	5.252521984\\
41	5.252452352\\
42	5.250383872\\
43	5.250277376\\
44	5.582352384\\
45	6.034878464\\
46	6.034755584\\
47	6.034755584\\
48	6.034735104\\
49	6.03475968\\
50	6.035427328\\
51	6.03531264\\
52	6.035222528\\
53	6.03506688\\
54	6.194245632\\
55	6.405419008\\
56	6.572937216\\
57	6.532804608\\
58	6.556229632\\
59	6.5939456\\
60	6.549131264\\
61	6.5968128\\
62	6.638211072\\
63	6.653431808\\
64	6.653431808\\
65	6.653374464\\
66	6.6533376\\
67	6.652522496\\
68	6.654230528\\
69	6.654459904\\
70	6.654418944\\
71	6.654418944\\
72	6.654418944\\
73	6.654418944\\
74	6.899654656\\
75	6.842310656\\
76	6.817222656\\
77	6.792482816\\
78	6.840623104\\
79	6.8728832\\
80	6.895271936\\
81	6.903291904\\
82	6.951141376\\
83	6.9900288\\
84	6.961463296\\
85	6.855593984\\
86	6.893780992\\
87	6.938476544\\
88	6.940393472\\
89	6.963154944\\
90	6.977806336\\
91	7.003295744\\
92	6.859845632\\
93	6.906056704\\
94	6.932406272\\
95	6.954405888\\
96	6.945153024\\
97	6.64475648\\
98	6.666596352\\
99	6.715736064\\
100	6.7666944\\
101	6.814662656\\
102	6.86016512\\
103	6.902890496\\
104	6.791200768\\
105	6.833938432\\
106	6.87157248\\
107	6.78111232\\
108	6.821933056\\
109	6.860439552\\
110	6.798893056\\
111	6.827163648\\
112	6.858293248\\
113	6.878502912\\
114	6.891012096\\
115	6.89094656\\
116	6.89094656\\
117	6.877872128\\
118	6.918762496\\
119	6.875254784\\
120	6.87478784\\
121	6.873473024\\
122	6.864203776\\
123	6.85989888\\
124	6.85989888\\
125	6.975033344\\
};
\addlegendentry{D$^*_+$}

\addplot[area legend, draw=black, fill=black!20!red, fill opacity=0.3]
table[row sep=crcr] {%
x	y\\
18	0\\
19	0\\
20	0\\
21	0\\
22	0\\
23	0\\
24	0\\
25	0\\
26	0\\
27	0\\
28	0\\
29	0\\
30	0\\
31	0\\
32	0\\
33	0\\
34	0\\
35	0\\
36	0\\
37	0\\
38	0\\
39	0\\
40	0\\
41	0\\
42	0\\
43	0\\
44	0\\
45	0\\
46	0\\
47	0\\
48	0\\
49	0\\
50	0\\
51	0\\
52	0\\
53	0\\
54	0\\
54	6.194245632\\
53	6.03506688\\
52	6.035222528\\
51	6.03531264\\
50	6.035427328\\
49	6.03475968\\
48	6.034735104\\
47	6.034755584\\
46	6.034755584\\
45	6.034878464\\
44	5.582352384\\
43	5.250277376\\
42	5.250383872\\
41	5.252452352\\
40	5.252521984\\
39	5.252542464\\
38	5.252579328\\
37	5.252653056\\
36	5.252653056\\
35	5.204533248\\
34	4.733210624\\
33	4.422647808\\
32	4.422647808\\
31	4.422647808\\
30	4.422647808\\
29	4.422647808\\
28	4.422647808\\
27	4.420214784\\
26	3.915767808\\
25	3.77843712\\
24	3.77843712\\
23	3.77843712\\
22	3.77843712\\
21	3.77843712\\
20	3.77843712\\
19	3.502964736\\
18	3.502964736\\
}--cycle;
\addlegendentry{D$^*_+$ $\mathbb{G}$ expansion}

\addplot[area legend, draw=black, fill=black!20!red, fill opacity=0.3, forget plot]
table[row sep=crcr] {%
x	y\\
65	0\\
66	0\\
67	0\\
68	0\\
69	0\\
70	0\\
71	0\\
72	0\\
73	0\\
74	0\\
74	6.899654656\\
73	6.654418944\\
72	6.654418944\\
71	6.654418944\\
70	6.654418944\\
69	6.654459904\\
68	6.654230528\\
67	6.652522496\\
66	6.6533376\\
65	6.653374464\\
64	6.653431808\\
}--cycle;

\addplot[area legend, draw=black, fill=black!20!red, fill opacity=0.3, forget plot]
table[row sep=crcr] {%
x	y\\
115	0\\
116	0\\
117	0\\
118	0\\
118	6.918762496\\
117	6.877872128\\
116	6.89094656\\
115	6.89094656\\
}--cycle;

\addplot[area legend, draw=black, fill=black!20!red, fill opacity=0.3, forget plot]
table[row sep=crcr] {%
x	y\\
120	0\\
121	0\\
122	0\\
123	0\\
124	0\\
125	0\\
125	6.975033344\\
124	6.85989888\\
123	6.85989888\\
122	6.864203776\\
121	6.873473024\\
120	6.87478784\\
}--cycle;
\end{axis}
\end{tikzpicture}%
    \caption{The memory consumption of D$^*_+$ and Gbplanner in the comparison experiment. The shaded areas mark the time when D$^*_+$ performed expansion of $\mathbb{G}$.}
    \label{fig:mem_comp}
\end{figure}
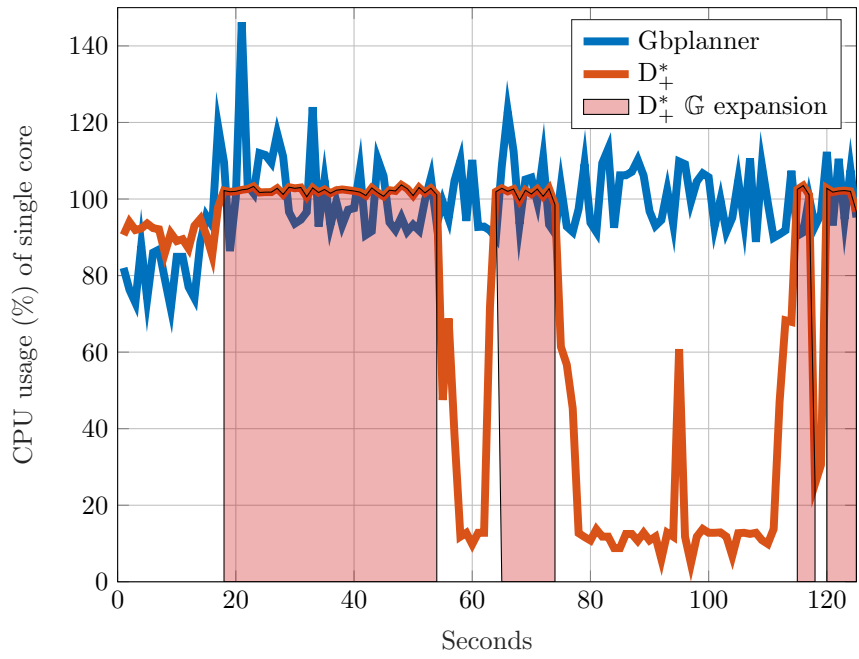
\begin{figure}
    \center
    \setlength\fwidth{0.9\columnwidth}
%
%
\definecolor{mycolor1}{rgb}{0.00000,0.44700,0.74100}%
\definecolor{mycolor2}{rgb}{0.85000,0.32500,0.09800}%
\begin{tikzpicture}

\begin{axis}[%
width=0.9\fwidth,
height=0.7\fwidth,
at={(0.9\fwidth,0.9\fwidth)},
scale only axis,
xmin=0,
xmax=125,
xlabel style={font=\color{white!15!black}},
xlabel={Seconds},
ymin=0,
ymax=150,
ylabel style={font=\color{white!15!black}},
ylabel={CPU usage (\%) of single core},
axis background/.style={fill=white},
xmajorgrids,
ymajorgrids,
legend style={legend cell align=left, align=left, draw=white!15!black}
]
\addplot [color=mycolor1, line width=3.0pt]
  table[row sep=crcr]{%
1	82\\
2	76.3000030517578\\
3	72.9000015258789\\
4	87.8000030517578\\
5	73.9000015258789\\
6	86\\
7	86.8000030517578\\
8	79\\
9	71.4000015258789\\
10	84.9000015258789\\
11	84.9000015258789\\
12	77\\
13	74.0999984741211\\
14	88.6999969482422\\
15	95.5999984741211\\
16	93.0999984741211\\
17	118.900001525879\\
18	109.599998474121\\
19	86.4000015258789\\
20	100.5\\
21	146.199996948242\\
22	103.5\\
23	101.5\\
24	112\\
25	111.5\\
26	109.5\\
27	117.099998474121\\
28	111.099998474121\\
29	96.5999984741211\\
30	93.5999984741211\\
31	94.5999984741211\\
32	96.5999984741211\\
33	124\\
34	92.8000030517578\\
35	105\\
36	93.0999984741211\\
37	99.3000030517578\\
38	93.1999969482422\\
39	97.1999969482422\\
40	97.5999984741211\\
41	107.599998474121\\
42	90.5999984741211\\
43	91.5999984741211\\
44	110.400001525879\\
45	106\\
46	93.9000015258789\\
47	91.9000015258789\\
48	95.1999969482422\\
49	91.5\\
50	93.3000030517578\\
51	91.8000030517578\\
52	99.6999969482422\\
53	105.599998474121\\
54	93.3000030517578\\
55	99.3000030517578\\
56	94.8000030517578\\
57	104.800003051758\\
58	110.099998474121\\
59	94.3000030517578\\
60	110.199996948242\\
61	92.6999969482422\\
62	92.8000030517578\\
63	91.8000030517578\\
64	89.8000030517578\\
65	108.599998474121\\
66	122.900001525879\\
67	112.400001525879\\
68	93.4000015258789\\
69	105.099998474121\\
70	105.599998474121\\
71	99.3000030517578\\
72	110.699996948242\\
73	93.3000030517578\\
74	91\\
75	103.5\\
76	92.8000030517578\\
77	91.4000015258789\\
78	97.4000015258789\\
79	109.199996948242\\
80	93.8000030517578\\
81	91.4000015258789\\
82	109.400001525879\\
83	113.099998474121\\
84	92.5\\
85	106.199996948242\\
86	106.199996948242\\
87	107.099998474121\\
88	110.199996948242\\
89	106.099998474121\\
90	96.8000030517578\\
91	93\\
92	94.5\\
93	102.300003051758\\
94	93.5\\
95	109.800003051758\\
96	109.099998474121\\
97	99.3000030517578\\
98	104.599998474121\\
99	106.699996948242\\
100	105.699996948242\\
101	92.8000030517578\\
102	99.8000030517578\\
103	91.5\\
104	95.0999984741211\\
105	104.900001525879\\
106	93.9000015258789\\
107	110.599998474121\\
108	88.8000030517578\\
109	108.900001525879\\
110	99.0999984741211\\
111	89.9000015258789\\
112	90.8000030517578\\
113	91.8000030517578\\
114	107.400001525879\\
115	90.4000015258789\\
116	91.3000030517578\\
117	100.400001525879\\
118	92.9000015258789\\
119	96.0999984741211\\
120	112.300003051758\\
121	93.0999984741211\\
122	110.5\\
123	94.6999969482422\\
124	107.599998474121\\
125	95.1999969482422\\
};
\addlegendentry{Gbplanner}

\addplot [color=mycolor2, line width=3.0pt]
  table[row sep=crcr]{%
1	90.5999984741211\\
2	94\\
3	91.9000015258789\\
4	92.3000030517578\\
5	93.5999984741211\\
6	92.4000015258789\\
7	92.0999984741211\\
8	85.9000015258789\\
9	91.0999984741211\\
10	89\\
11	89.5\\
12	87.0999984741211\\
13	93\\
14	94.8000030517578\\
15	90.5999984741211\\
16	85.3000030517578\\
17	97.9000015258789\\
18	102.199996948242\\
19	101.900001525879\\
20	102\\
21	102.400001525879\\
22	102.599998474121\\
23	103.300003051758\\
24	101.699996948242\\
25	101.800003051758\\
26	101.800003051758\\
27	102.800003051758\\
28	101.199996948242\\
29	103.099998474121\\
30	102.800003051758\\
31	103\\
32	100.599998474121\\
33	102.900001525879\\
34	101.699996948242\\
35	102.5\\
36	101.5\\
37	102.300003051758\\
38	102.5\\
39	102.300003051758\\
40	102.099998474121\\
41	101.800003051758\\
42	100.900001525879\\
43	102.900001525879\\
44	101.699996948242\\
45	100.699996948242\\
46	102.300003051758\\
47	102.099998474121\\
48	103.699996948242\\
49	102.699996948242\\
50	100.900001525879\\
51	103\\
52	101.599998474121\\
53	102.800003051758\\
54	101.099998474121\\
55	47.5\\
56	68.8000030517578\\
57	38.2999992370605\\
58	11.8000001907349\\
59	12.8000001907349\\
60	9.80000019073486\\
61	12.8000001907349\\
62	12.8000001907349\\
63	71.8000030517578\\
64	101.900001525879\\
65	102.800003051758\\
66	102\\
67	102.599998474121\\
68	99.5999984741211\\
69	102.300003051758\\
70	101.300003051758\\
71	102.599998474121\\
72	100.699996948242\\
73	102.699996948242\\
74	98.5\\
75	61.4000015258789\\
76	56.7000007629395\\
77	45.2999992370605\\
78	12.6999998092651\\
79	11.6999998092651\\
80	10.8000001907349\\
81	13.6999998092651\\
82	11.8000001907349\\
83	11.8000001907349\\
84	8.80000019073486\\
85	8.80000019073486\\
86	12.5\\
87	12.5\\
88	10.6999998092651\\
89	12.8000001907349\\
90	10.8000001907349\\
91	11.8000001907349\\
92	5.90000009536743\\
93	12.8000001907349\\
94	11.8000001907349\\
95	60.7999992370605\\
96	11.8000001907349\\
97	4.80000019073486\\
98	11.8000001907349\\
99	13.8000001907349\\
100	12.8000001907349\\
101	12.8000001907349\\
102	12.8999996185303\\
103	11.8000001907349\\
104	6.80000019073486\\
105	12.6999998092651\\
106	12.8000001907349\\
107	12.5\\
108	12.8000001907349\\
109	10.8999996185303\\
110	9.89999961853027\\
111	13.8000001907349\\
112	47\\
113	68.3000030517578\\
114	67.9000015258789\\
115	102.5\\
116	103.5\\
117	101.099998474121\\
118	25.6000003814697\\
119	30.5\\
120	102.800003051758\\
121	101.800003051758\\
122	102.099998474121\\
123	102.199996948242\\
124	102\\
125	96.6999969482422\\
};
\addlegendentry{D$^*_+$}

\addplot[area legend, draw=black, fill=black!20!red, fill opacity=0.3]
table[row sep=crcr] {%
x	y\\
18	0\\
19	0\\
20	0\\
21	0\\
22	0\\
23	0\\
24	0\\
25	0\\
26	0\\
27	0\\
28	0\\
29	0\\
30	0\\
31	0\\
32	0\\
33	0\\
34	0\\
35	0\\
36	0\\
37	0\\
38	0\\
39	0\\
40	0\\
41	0\\
42	0\\
43	0\\
44	0\\
45	0\\
46	0\\
47	0\\
48	0\\
49	0\\
50	0\\
51	0\\
52	0\\
53	0\\
54	0\\
54	101.099998474121\\
53	102.800003051758\\
52	101.599998474121\\
51	103\\
50	100.900001525879\\
49	102.699996948242\\
48	103.699996948242\\
47	102.099998474121\\
46	102.300003051758\\
45	100.699996948242\\
44	101.699996948242\\
43	102.900001525879\\
42	100.900001525879\\
41	101.800003051758\\
40	102.099998474121\\
39	102.300003051758\\
38	102.5\\
37	102.300003051758\\
36	101.5\\
35	102.5\\
34	101.699996948242\\
33	102.900001525879\\
32	100.599998474121\\
31	103\\
30	102.800003051758\\
29	103.099998474121\\
28	101.199996948242\\
27	102.800003051758\\
26	101.800003051758\\
25	101.800003051758\\
24	101.699996948242\\
23	103.300003051758\\
22	102.599998474121\\
21	102.400001525879\\
20	102\\
19	101.900001525879\\
18	102.199996948242\\
}--cycle;
\addlegendentry{D$^*_+$ $\mathbb{G}$ expansion}

\addplot[area legend, draw=black, fill=black!20!red, fill opacity=0.3, forget plot]
table[row sep=crcr] {%
x	y\\
65	0\\
66	0\\
67	0\\
68	0\\
69	0\\
70	0\\
71	0\\
72	0\\
73	0\\
74	0\\
74	98.5\\
73	102.699996948242\\
72	100.699996948242\\
71	102.599998474121\\
70	101.300003051758\\
69	102.300003051758\\
68	99.5999984741211\\
67	102.599998474121\\
66	102\\
65	102.800003051758\\
64	101.900001525879\\
}--cycle;

\addplot[area legend, draw=black, fill=black!20!red, fill opacity=0.3, forget plot]
table[row sep=crcr] {%
x	y\\
115	0\\
116	0\\
117	0\\
118	0\\
118	25.6000003814697\\
117	101.099998474121\\
116	103.5\\
115	102.5\\
}--cycle;

\addplot[area legend, draw=black, fill=black!20!red, fill opacity=0.3, forget plot]
table[row sep=crcr] {%
x	y\\
120	0\\
121	0\\
122	0\\
123	0\\
124	0\\
125	0\\
125	96.6999969482422\\
124	102\\
123	102.199996948242\\
122	102.099998474121\\
121	101.800003051758\\
120	102.800003051758\\
}--cycle;
\end{axis}
\end{tikzpicture}%
    \caption{The CPU load comparison of D$^*_+$ and Gbplanner in the comparison experiment. The shaded areas mark the time when D$^*_+$ performed expansion of $\mathbb{G}$.}
    \label{fig:cpu_comp}
\end{figure}
D$^*_+$ capability to replan a path on the fly can not be compared to Gbplanner because it does not provide an equivalent feature. Some of the differences in performance can be explained by differences in those types of features that in some cases affects the measurable performance.

\section{Conclusions}\label{sec:con}
In this article, a risk-aware, platform-agnostic path planner called D$^*_+$ has been established. As it was presented, D$^*_+$ utilizes a proximity risk layer to generate a safety margin for any occupied space. The introduced combination of proximity risk and unknown space treatment is able to solve the common issue of shortcuts due to imperfections in the map. By allowing the map to update and expand, the proposed framework is capable of exploring and operating in a previously unknown environment.
Finally, the efficiency of the D$^*_+$ has been tested in challenging real-world scenarios with both 2D and 3D planning and it has been proven to reliably plan safe paths in all the evaluated use cases independently of the utilized platform.

Considering large-scale and long-term missions, a potential bottleneck of D$^*_+$ is its computation time when $\mathbf{G}$ is large, and as such future work should aim to improve the computation time for large $\mathbf{G}$ by considering for example map segmentation approaches. Additionally, another point of D$^*_+$ that should be addressed is the severe memory usage. The comparison with Gbplanner showed that it is possible to do memory-efficient path planning that does not require more computation power. This part needs to be investigated and addressed for  D$^*_+$ while maintaining the major components of the short straight paths and the dynamic re-planning. The above-mentioned shortcoming in general is related to the resources required to compute $P$. One direction for lowering memory usage and possibly saving some computation time is to change the data structure of the internal gird map $\mathbf{G}$.
Currently, $\mathbf{G}$ is stored as an array in D$^*_+$ but if it is changed to an oct-tree it is possible that memory usage and computation time can be lowered.
Another improvement that can be considered is more types of risks, such as wind or slippery surfaces.

\bibliography{sample}

\begin{thebibliography}{38}
\expandafter\ifx\csname natexlab\endcsname\relax\def\natexlab#1{#1}\fi
\providecommand{\url}[1]{\texttt{#1}}
\providecommand{\href}[2]{#2}
\providecommand{\path}[1]{#1}
\providecommand{\DOIprefix}{doi:}
\providecommand{\ArXivprefix}{arXiv:}
\providecommand{\URLprefix}{URL: }
\providecommand{\Pubmedprefix}{pmid:}
\providecommand{\doi}[1]{\href{http://dx.doi.org/#1}{\path{#1}}}
\providecommand{\Pubmed}[1]{\href{pmid:#1}{\path{#1}}}
\providecommand{\bibinfo}[2]{#2}
\ifx\xfnm\relax \def\xfnm[#1]{\unskip,\space#1}\fi
\bibitem[{Agha et~al.(2021)Agha, Otsu, Morrell, Fan, Thakker,
  Santamaria-Navarro, Kim, Bouman, Lei, Edlund et~al.}]{agha2021nebula}
\bibinfo{author}{Agha, A.}, \bibinfo{author}{Otsu, K.},
  \bibinfo{author}{Morrell, B.}, \bibinfo{author}{Fan, D.~D.},
  \bibinfo{author}{Thakker, R.}, \bibinfo{author}{Santamaria-Navarro, A.},
  \bibinfo{author}{Kim, S.-K.}, \bibinfo{author}{Bouman, A.},
  \bibinfo{author}{Lei, X.}, \bibinfo{author}{Edlund, J.} et~al.
  (\bibinfo{year}{2021}).
\newblock \bibinfo{title}{Nebula: Quest for robotic autonomy in challenging
  environments; team costar at the darpa subterranean challenge}.
\newblock {\it \bibinfo{journal}{arXiv preprint arXiv:2103.11470}\/}, .
\bibitem[{Atapour-Abarghouei \& Breckon(2018)}]{ATAPOURABARGHOUEI201839}
\bibinfo{author}{Atapour-Abarghouei, A.}, \& \bibinfo{author}{Breckon, T.~P.}
  (\bibinfo{year}{2018}).
\newblock \bibinfo{title}{A comparative review of plausible hole filling
  strategies in the context of scene depth image completion}.
\newblock {\it \bibinfo{journal}{Computers \& Graphics}\/},  {\it
  \bibinfo{volume}{72}\/}, \bibinfo{pages}{39--58}. \URLprefix
  \url{https://www.sciencedirect.com/science/article/pii/S0097849318300219}.
  \DOIprefix\doi{https://doi.org/10.1016/j.cag.2018.02.001}.
\bibitem[{Cuevas et~al.(2021)Cuevas, Ramirez, Shames \& Manzic}]{9483405}
\bibinfo{author}{Cuevas, L.}, \bibinfo{author}{Ramirez, M.},
  \bibinfo{author}{Shames, I.}, \& \bibinfo{author}{Manzic, C.}
  (\bibinfo{year}{2021}).
\newblock \bibinfo{title}{Path planning under risk and uncertainty of the
  environment}.
\newblock In {\it \bibinfo{booktitle}{2021 American Control Conference
  (ACC)}\/} (pp. \bibinfo{pages}{4231--4236}).
\newblock \DOIprefix\doi{10.23919/ACC50511.2021.9483405}.
\bibitem[{Dang et~al.(2020)Dang, Tranzatto, Khattak, Mascarich, Alexis \&
  Hutter}]{dang2020graph}
\bibinfo{author}{Dang, T.}, \bibinfo{author}{Tranzatto, M.},
  \bibinfo{author}{Khattak, S.}, \bibinfo{author}{Mascarich, F.},
  \bibinfo{author}{Alexis, K.}, \& \bibinfo{author}{Hutter, M.}
  (\bibinfo{year}{2020}).
\newblock \bibinfo{title}{Graph-based subterranean exploration path planning
  using aerial and legged robots}.
\newblock {\it \bibinfo{journal}{Journal of Field Robotics}\/},  {\it
  \bibinfo{volume}{37}\/}, \bibinfo{pages}{1363--1388}.
\newblock \bibinfo{note}{Wiley Online Library}.
\bibitem[{DARPA(2020)}]{subtworld}
\bibinfo{author}{DARPA} (\bibinfo{year}{2020}).
\newblock \bibinfo{title}{{DARPA} {S}ubterranean ({S}ub{T}) challenge}.
\newblock \URLprefix \url{https://www.subtchallenge.com/}
  \bibinfo{note}{accessed: February 2021}.
\bibitem[{Hakobyan et~al.(2019)Hakobyan, Kim \& Yang}]{hakobyan2019risk}
\bibinfo{author}{Hakobyan, A.}, \bibinfo{author}{Kim, G.~C.}, \&
  \bibinfo{author}{Yang, I.} (\bibinfo{year}{2019}).
\newblock \bibinfo{title}{Risk-aware motion planning and control using
  cvar-constrained optimization}.
\newblock {\it \bibinfo{journal}{IEEE Robotics and Automation letters}\/},
  {\it \bibinfo{volume}{4}\/}, \bibinfo{pages}{3924--3931}.
\bibitem[{Hayat et~al.(2020)Hayat, Yanmaz, Bettstetter \& Brown}]{Hayat2020}
\bibinfo{author}{Hayat, S.}, \bibinfo{author}{Yanmaz, E.},
  \bibinfo{author}{Bettstetter, C.}, \& \bibinfo{author}{Brown, T.~X.}
  (\bibinfo{year}{2020}).
\newblock \bibinfo{title}{Multi-objective drone path planning for search and
  rescue with quality-of-service requirements}.
\newblock {\it \bibinfo{journal}{Autonomous Robots}\/},  {\it
  \bibinfo{volume}{44}\/}, \bibinfo{pages}{1183--1198}. \URLprefix
  \url{https://doi.org/10.1007/s10514-020-09926-9}.
  \DOIprefix\doi{10.1007/s10514-020-09926-9}.
\bibitem[{Hess et~al.(2016)Hess, Kohler, Rapp \& Andor}]{45466}
\bibinfo{author}{Hess, W.}, \bibinfo{author}{Kohler, D.},
  \bibinfo{author}{Rapp, H.}, \& \bibinfo{author}{Andor, D.}
  (\bibinfo{year}{2016}).
\newblock \bibinfo{title}{Real-time loop closure in 2d lidar slam}.
\newblock In {\it \bibinfo{booktitle}{2016 IEEE International Conference on
  Robotics and Automation (ICRA)}\/} (pp. \bibinfo{pages}{1271--1278}).
\bibitem[{Hornung et~al.(2013)Hornung, Wurm, Bennewitz, Stachniss \&
  Burgard}]{hornung13auro}
\bibinfo{author}{Hornung, A.}, \bibinfo{author}{Wurm, K.~M.},
  \bibinfo{author}{Bennewitz, M.}, \bibinfo{author}{Stachniss, C.}, \&
  \bibinfo{author}{Burgard, W.} (\bibinfo{year}{2013}).
\newblock \bibinfo{title}{{OctoMap}: An efficient probabilistic {3D} mapping
  framework based on octrees}.
\newblock {\it \bibinfo{journal}{Autonomous Robots}\/}, . \URLprefix
  \url{https://octomap.github.io}. \DOIprefix\doi{10.1007/s10514-012-9321-0}.
\newblock \bibinfo{note}{Software available at
  \url{https://octomap.github.io}}.
\bibitem[{Hu et~al.(2020)Hu, Pang, Dai \& Low}]{9165709}
\bibinfo{author}{Hu, X.}, \bibinfo{author}{Pang, B.}, \bibinfo{author}{Dai,
  F.}, \& \bibinfo{author}{Low, K.~H.} (\bibinfo{year}{2020}).
\newblock \bibinfo{title}{Risk assessment model for uav cost-effective path
  planning in urban environments}.
\newblock {\it \bibinfo{journal}{IEEE Access}\/},  {\it \bibinfo{volume}{8}\/},
  \bibinfo{pages}{150162--150173}. \DOIprefix\doi{10.1109/ACCESS.2020.3016118}.
\bibitem[{Huang et~al.(2020)Huang, Schwarting, Pierson, Guo, Ang \&
  Rus}]{9341084}
\bibinfo{author}{Huang, Z.}, \bibinfo{author}{Schwarting, W.},
  \bibinfo{author}{Pierson, A.}, \bibinfo{author}{Guo, H.},
  \bibinfo{author}{Ang, M.}, \& \bibinfo{author}{Rus, D.}
  (\bibinfo{year}{2020}).
\newblock \bibinfo{title}{Safe path planning with multi-model risk level sets}.
\newblock In {\it \bibinfo{booktitle}{2020 IEEE/RSJ International Conference on
  Intelligent Robots and Systems (IROS)}\/} (pp. \bibinfo{pages}{6268--6275}).
\newblock \DOIprefix\doi{10.1109/IROS45743.2020.9341084}.
\bibitem[{Koenig \& Likhachev(2002)}]{koenig2002d}
\bibinfo{author}{Koenig, S.}, \& \bibinfo{author}{Likhachev, M.}
  (\bibinfo{year}{2002}).
\newblock \bibinfo{title}{D\*{*} lite}.
\newblock {\it \bibinfo{journal}{Aaai/iaai}\/},  {\it \bibinfo{volume}{15}\/}.
\bibitem[{Koval et~al.(2022)Koval, Karlsson \&
  Nikolakopoulos}]{koval2022experimental}
\bibinfo{author}{Koval, A.}, \bibinfo{author}{Karlsson, S.}, \&
  \bibinfo{author}{Nikolakopoulos, G.} (\bibinfo{year}{2022}).
\newblock \bibinfo{title}{Experimental evaluation of autonomous map-based spot
  navigation in confined environments}.
\newblock {\it \bibinfo{journal}{Biomimetic Intelligence and Robotics}\/},  (p.
  \bibinfo{pages}{100035}).
\bibitem[{Kulkarni et~al.(2022)Kulkarni, Dharmadhikari, Tranzatto, Zimmermann,
  Reijgwart, De~Petris, Nguyen, Khedekar, Papachristos, Ott, Siegwart, Hutter
  \& Alexis}]{9812401}
\bibinfo{author}{Kulkarni, M.}, \bibinfo{author}{Dharmadhikari, M.},
  \bibinfo{author}{Tranzatto, M.}, \bibinfo{author}{Zimmermann, S.},
  \bibinfo{author}{Reijgwart, V.}, \bibinfo{author}{De~Petris, P.},
  \bibinfo{author}{Nguyen, H.}, \bibinfo{author}{Khedekar, N.},
  \bibinfo{author}{Papachristos, C.}, \bibinfo{author}{Ott, L.},
  \bibinfo{author}{Siegwart, R.}, \bibinfo{author}{Hutter, M.}, \&
  \bibinfo{author}{Alexis, K.} (\bibinfo{year}{2022}).
\newblock \bibinfo{title}{Autonomous teamed exploration of subterranean
  environments using legged and aerial robots}.
\newblock In {\it \bibinfo{booktitle}{2022 International Conference on Robotics
  and Automation (ICRA)}\/} (pp. \bibinfo{pages}{3306--3313}).
\newblock \DOIprefix\doi{10.1109/ICRA46639.2022.9812401}.
\bibitem[{Laconte et~al.(2021)Laconte, Kasmi, Pomerleau, Chapuis, Malaterre,
  Debain \& Aufr{\`e}re}]{laconte2021novel}
\bibinfo{author}{Laconte, J.}, \bibinfo{author}{Kasmi, A.},
  \bibinfo{author}{Pomerleau, F.}, \bibinfo{author}{Chapuis, R.},
  \bibinfo{author}{Malaterre, L.}, \bibinfo{author}{Debain, C.}, \&
  \bibinfo{author}{Aufr{\`e}re, R.} (\bibinfo{year}{2021}).
\newblock \bibinfo{title}{A novel occupancy mapping framework for risk-aware
  path planning in unstructured environments}.
\newblock {\it \bibinfo{journal}{Sensors}\/},  {\it \bibinfo{volume}{21}\/},
  \bibinfo{pages}{7562}.
\bibitem[{Li et~al.(2018)Li, Zlatanova, Koopman, Bai \& Diakité}]{LI2018275}
\bibinfo{author}{Li, F.}, \bibinfo{author}{Zlatanova, S.},
  \bibinfo{author}{Koopman, M.}, \bibinfo{author}{Bai, X.}, \&
  \bibinfo{author}{Diakité, A.} (\bibinfo{year}{2018}).
\newblock \bibinfo{title}{Universal path planning for an indoor drone}.
\newblock {\it \bibinfo{journal}{Automation in Construction}\/},  {\it
  \bibinfo{volume}{95}\/}, \bibinfo{pages}{275--283}. \URLprefix
  \url{https://www.sciencedirect.com/science/article/pii/S0926580517311184}.
  \DOIprefix\doi{https://doi.org/10.1016/j.autcon.2018.07.025}.
\bibitem[{Lindqvist et~al.(2021{\natexlab{a}})Lindqvist, Kanellakis, Mansouri,
  akbar Agha-mohammadi \& Nikolakopoulos}]{lindqvist2021compra}
\bibinfo{author}{Lindqvist, B.}, \bibinfo{author}{Kanellakis, C.},
  \bibinfo{author}{Mansouri, S.~S.}, \bibinfo{author}{akbar Agha-mohammadi,
  A.}, \& \bibinfo{author}{Nikolakopoulos, G.}
  (\bibinfo{year}{2021}{\natexlab{a}}).
\newblock \bibinfo{title}{Compra: A compact reactive autonomy framework for
  subterranean mav based search-and-rescue operations}.
\newblock \href{http://arxiv.org/abs/2108.13105}{\tt arXiv:2108.13105}.
\bibitem[{Lindqvist et~al.(2021{\natexlab{b}})Lindqvist, Mansouri, Haluška \&
  Nikolakopoulos}]{9625659}
\bibinfo{author}{Lindqvist, B.}, \bibinfo{author}{Mansouri, S.~S.},
  \bibinfo{author}{Haluška, J.}, \& \bibinfo{author}{Nikolakopoulos, G.}
  (\bibinfo{year}{2021}{\natexlab{b}}).
\newblock \bibinfo{title}{Reactive navigation of an unmanned aerial vehicle
  with perception-based obstacle avoidance constraints}.
\newblock {\it \bibinfo{journal}{IEEE Transactions on Control Systems
  Technology}\/},  (pp. \bibinfo{pages}{1--16}).
  \DOIprefix\doi{10.1109/TCST.2021.3124820}.
\bibitem[{Mishra et~al.(2020)Mishra, Garg, Narang \& Mishra}]{MISHRA20201}
\bibinfo{author}{Mishra, B.}, \bibinfo{author}{Garg, D.},
  \bibinfo{author}{Narang, P.}, \& \bibinfo{author}{Mishra, V.}
  (\bibinfo{year}{2020}).
\newblock \bibinfo{title}{Drone-surveillance for search and rescue in natural
  disaster}.
\newblock {\it \bibinfo{journal}{Computer Communications}\/},  {\it
  \bibinfo{volume}{156}\/}, \bibinfo{pages}{1--10}. \URLprefix
  \url{https://www.sciencedirect.com/science/article/pii/S0140366419318602}.
  \DOIprefix\doi{https://doi.org/10.1016/j.comcom.2020.03.012}.
\bibitem[{Ono et~al.(2015)Ono, Fuchs, Steffy, Maimone \& Yen}]{7119022}
\bibinfo{author}{Ono, M.}, \bibinfo{author}{Fuchs, T.~J.},
  \bibinfo{author}{Steffy, A.}, \bibinfo{author}{Maimone, M.}, \&
  \bibinfo{author}{Yen, J.} (\bibinfo{year}{2015}).
\newblock \bibinfo{title}{Risk-aware planetary rover operation: Autonomous
  terrain classification and path planning}.
\newblock In {\it \bibinfo{booktitle}{2015 IEEE Aerospace Conference}\/} (pp.
  \bibinfo{pages}{1--10}).
\newblock \DOIprefix\doi{10.1109/AERO.2015.7119022}.
\bibitem[{Primatesta et~al.(2017)Primatesta, Capello, Antonini, Gaspardone,
  Guglieri \& Rizzo}]{7991358}
\bibinfo{author}{Primatesta, S.}, \bibinfo{author}{Capello, E.},
  \bibinfo{author}{Antonini, R.}, \bibinfo{author}{Gaspardone, M.},
  \bibinfo{author}{Guglieri, G.}, \& \bibinfo{author}{Rizzo, A.}
  (\bibinfo{year}{2017}).
\newblock \bibinfo{title}{A cloud-based framework for risk-aware intelligent
  navigation in urban environments}.
\newblock In {\it \bibinfo{booktitle}{2017 International Conference on Unmanned
  Aircraft Systems (ICUAS)}\/} (pp. \bibinfo{pages}{447--455}).
\newblock \DOIprefix\doi{10.1109/ICUAS.2017.7991358}.
\bibitem[{Primatesta et~al.(2019)Primatesta, Guglieri \&
  Rizzo}]{Primatesta2019}
\bibinfo{author}{Primatesta, S.}, \bibinfo{author}{Guglieri, G.}, \&
  \bibinfo{author}{Rizzo, A.} (\bibinfo{year}{2019}).
\newblock \bibinfo{title}{A risk-aware path planning strategy for uavs in urban
  environments}.
\newblock {\it \bibinfo{journal}{Journal of Intelligent {\&} Robotic
  Systems}\/},  {\it \bibinfo{volume}{95}\/}, \bibinfo{pages}{629--643}.
  \URLprefix \url{https://doi.org/10.1007/s10846-018-0924-3}.
  \DOIprefix\doi{10.1007/s10846-018-0924-3}.
\bibitem[{Puck et~al.(2020)Puck, Schnell, Plasberg, B{\"u}ttner, Heppner,
  R{\"o}nnau \& Dillmann}]{puck2020modular}
\bibinfo{author}{Puck, L.}, \bibinfo{author}{Schnell, T.},
  \bibinfo{author}{Plasberg, C.}, \bibinfo{author}{B{\"u}ttner, T.},
  \bibinfo{author}{Heppner, G.}, \bibinfo{author}{R{\"o}nnau, A.}, \&
  \bibinfo{author}{Dillmann, R.} (\bibinfo{year}{2020}).
\newblock \bibinfo{title}{Modular, risk-aware mapping and fusion of
  environmental hazards}.
\newblock In {\it \bibinfo{booktitle}{2020 IEEE 23rd International Conference
  on Information Fusion (FUSION)}\/} (pp. \bibinfo{pages}{1--6}).
\newblock \bibinfo{organization}{IEEE}.
\bibitem[{San~Juan et~al.(2018)San~Juan, Santos \& And{\'u}jar}]{SanJuan2018}
\bibinfo{author}{San~Juan, V.}, \bibinfo{author}{Santos, M.}, \&
  \bibinfo{author}{And{\'u}jar, J.~M.} (\bibinfo{year}{2018}).
\newblock \bibinfo{title}{Intelligent uav map generation and discrete path
  planning for search and rescue operations}.
\newblock {\it \bibinfo{journal}{Complexity}\/},  {\it
  \bibinfo{volume}{2018}\/}, \bibinfo{pages}{6879419}. \URLprefix
  \url{https://doi.org/10.1155/2018/6879419}.
  \DOIprefix\doi{10.1155/2018/6879419}.
\bibitem[{Schedl et~al.(2021)Schedl, Kurmi \&
  Bimber}]{doi:10.1126/scirobotics.abg1188}
\bibinfo{author}{Schedl, D.~C.}, \bibinfo{author}{Kurmi, I.}, \&
  \bibinfo{author}{Bimber, O.} (\bibinfo{year}{2021}).
\newblock \bibinfo{title}{An autonomous drone for search and rescue in forests
  using airborne optical sectioning}.
\newblock {\it \bibinfo{journal}{Science Robotics}\/},  {\it
  \bibinfo{volume}{6}\/}, \bibinfo{pages}{eabg1188}. \URLprefix
  \url{https://www.science.org/doi/abs/10.1126/scirobotics.abg1188}.
  \DOIprefix\doi{10.1126/scirobotics.abg1188}.
  \href{http://arxiv.org/abs/https://www.science.org/doi/pdf/10.1126/scirobotics.abg1188}{\tt
  arXiv:https://www.science.org/doi/pdf/10.1126/scirobotics.abg1188}.
\bibitem[{Shan et~al.(2020)Shan, Englot, Meyers, Wang, Ratti \&
  Daniela}]{liosam2020shan}
\bibinfo{author}{Shan, T.}, \bibinfo{author}{Englot, B.},
  \bibinfo{author}{Meyers, D.}, \bibinfo{author}{Wang, W.},
  \bibinfo{author}{Ratti, C.}, \& \bibinfo{author}{Daniela, R.}
  (\bibinfo{year}{2020}).
\newblock \bibinfo{title}{Lio-sam: Tightly-coupled lidar inertial odometry via
  smoothing and mapping}.
\newblock In {\it \bibinfo{booktitle}{IEEE/RSJ International Conference on
  Intelligent Robots and Systems (IROS)}\/} (pp. \bibinfo{pages}{5135--5142}).
\newblock \bibinfo{organization}{IEEE}.
\bibitem[{She \& Ouyang(2021)}]{SHE2021102878}
\bibinfo{author}{She, R.}, \& \bibinfo{author}{Ouyang, Y.}
  (\bibinfo{year}{2021}).
\newblock \bibinfo{title}{Efficiency of uav-based last-mile delivery under
  congestion in low-altitude air}.
\newblock {\it \bibinfo{journal}{Transportation Research Part C: Emerging
  Technologies}\/},  {\it \bibinfo{volume}{122}\/}, \bibinfo{pages}{102878}.
  \URLprefix
  \url{https://www.sciencedirect.com/science/article/pii/S0968090X20307786}.
  \DOIprefix\doi{https://doi.org/10.1016/j.trc.2020.102878}.
\bibitem[{da~Silva~Arantes et~al.(2019)da~Silva~Arantes, Toledo, Williams \&
  Ono}]{da2019collision}
\bibinfo{author}{da~Silva~Arantes, M.}, \bibinfo{author}{Toledo, C. F.~M.},
  \bibinfo{author}{Williams, B.~C.}, \& \bibinfo{author}{Ono, M.}
  (\bibinfo{year}{2019}).
\newblock \bibinfo{title}{Collision-free encoding for chance-constrained
  nonconvex path planning}.
\newblock {\it \bibinfo{journal}{IEEE Transactions on Robotics}\/},  {\it
  \bibinfo{volume}{35}\/}, \bibinfo{pages}{433--448}.
\bibitem[{{Stanford Artificial Intelligence Laboratory et al.}()}]{ros}
\bibinfo{author}{{Stanford Artificial Intelligence Laboratory et al.}} ().
\newblock \bibinfo{title}{Robotic operating system}.
\newblock \URLprefix \url{https://www.ros.org}.
\bibitem[{Tordesillas \& How(2021)}]{tordesillas2021faster}
\bibinfo{author}{Tordesillas, J.}, \& \bibinfo{author}{How, J.~P.}
  (\bibinfo{year}{2021}).
\newblock \bibinfo{title}{{FASTER}: Fast and safe trajectory planner for
  navigation in unknown environments}.
\newblock {\it \bibinfo{journal}{IEEE Transactions on Robotics}\/}, .
\bibitem[{Tordesillas et~al.(2019)Tordesillas, Lopez \&
  How}]{tordesillas2019faster}
\bibinfo{author}{Tordesillas, J.}, \bibinfo{author}{Lopez, B.~T.}, \&
  \bibinfo{author}{How, J.~P.} (\bibinfo{year}{2019}).
\newblock \bibinfo{title}{{FASTER}: Fast and safe trajectory planner for
  flights in unknown environments}.
\newblock In {\it \bibinfo{booktitle}{2019 IEEE/RSJ International Conference on
  Intelligent Robots and Systems (IROS)}\/}.
\newblock \bibinfo{organization}{IEEE}.
\bibitem[{Wang et~al.(2021)Wang, Ye, Wang, Gao, Xu \&
  Gao}]{wang2021learningbased}
\bibinfo{author}{Wang, L.}, \bibinfo{author}{Ye, H.}, \bibinfo{author}{Wang,
  Q.}, \bibinfo{author}{Gao, Y.}, \bibinfo{author}{Xu, C.}, \&
  \bibinfo{author}{Gao, F.} (\bibinfo{year}{2021}).
\newblock \bibinfo{title}{Learning-based 3d occupancy prediction for autonomous
  navigation in occluded environments}.
\newblock \href{http://arxiv.org/abs/2011.03981}{\tt arXiv:2011.03981}.
\bibitem[{Wu et~al.(2019)Wu, Xu, Zhen \& Wu}]{Wu2019}
\bibinfo{author}{Wu, X.}, \bibinfo{author}{Xu, L.}, \bibinfo{author}{Zhen, R.},
  \& \bibinfo{author}{Wu, X.} (\bibinfo{year}{2019}).
\newblock \bibinfo{title}{Biased sampling potentially guided intelligent
  bidirectional rrt<sup>∗</sup> algorithm for uav path planning in 3d
  environment}.
\newblock {\it \bibinfo{journal}{Mathematical Problems in Engineering}\/},
  {\it \bibinfo{volume}{2019}\/}, \bibinfo{pages}{5157403}. \URLprefix
  \url{https://doi.org/10.1155/2019/5157403}.
  \DOIprefix\doi{10.1155/2019/5157403}.
\bibitem[{Xiong et~al.(2021)Xiong, Gao, Wang, Li \& Lin}]{9367235}
\bibinfo{author}{Xiong, J.}, \bibinfo{author}{Gao, H.}, \bibinfo{author}{Wang,
  M.}, \bibinfo{author}{Li, H.}, \& \bibinfo{author}{Lin, W.}
  (\bibinfo{year}{2021}).
\newblock \bibinfo{title}{Occupancy map guided fast video-based dynamic point
  cloud coding}.
\newblock {\it \bibinfo{journal}{IEEE Transactions on Circuits and Systems for
  Video Technology}\/},  (pp. \bibinfo{pages}{1--1}).
  \DOIprefix\doi{10.1109/TCSVT.2021.3063501}.
\bibitem[{Yan~F.(2013)}]{Yan2013}
\bibinfo{author}{Yan~F., . X.~J., Liu~YS.} (\bibinfo{year}{2013}).
\newblock \bibinfo{title}{Path planning in complex 3d environments using a
  probabilistic roadmap method}.
\newblock {\it \bibinfo{journal}{International Journal of Automation and
  Computing}\/},  {\it \bibinfo{volume}{10}\/}, \bibinfo{pages}{525--533}.
  \DOIprefix\doi{https://doi.org/10.1007/s11633-013-0750-9}.
\bibitem[{Zammit \& Kampen()}]{doi:10.2514/6.2018-1846}
\bibinfo{author}{Zammit, C.}, \& \bibinfo{author}{Kampen, E.-J.~V.} ().
\newblock \bibinfo{title}{Comparison between a* and rrt algorithms for uav path
  planning}.
\newblock In {\it \bibinfo{booktitle}{2018 AIAA Guidance, Navigation, and
  Control Conference}\/}.
\newblock \URLprefix \url{https://arc.aiaa.org/doi/abs/10.2514/6.2018-1846}.
  \DOIprefix\doi{10.2514/6.2018-1846}.
  \href{http://arxiv.org/abs/https://arc.aiaa.org/doi/pdf/10.2514/6.2018-1846}{\tt
  arXiv:https://arc.aiaa.org/doi/pdf/10.2514/6.2018-1846}.
\bibitem[{Zhang et~al.(2021)Zhang, Zhang \& Low}]{ZHANG2021103123}
\bibinfo{author}{Zhang, N.}, \bibinfo{author}{Zhang, M.}, \&
  \bibinfo{author}{Low, K.~H.} (\bibinfo{year}{2021}).
\newblock \bibinfo{title}{3d path planning and real-time collision resolution
  of multirotor drone operations in complex urban low-altitude airspace}.
\newblock {\it \bibinfo{journal}{Transportation Research Part C: Emerging
  Technologies}\/},  {\it \bibinfo{volume}{129}\/}, \bibinfo{pages}{103123}.
  \URLprefix
  \url{https://www.sciencedirect.com/science/article/pii/S0968090X2100142X}.
  \DOIprefix\doi{https://doi.org/10.1016/j.trc.2021.103123}.
\bibitem[{Zhou et~al.(2021)Zhou, Pan, Gao \& Shen}]{zhou2021raptor}
\bibinfo{author}{Zhou, B.}, \bibinfo{author}{Pan, J.}, \bibinfo{author}{Gao,
  F.}, \& \bibinfo{author}{Shen, S.} (\bibinfo{year}{2021}).
\newblock \bibinfo{title}{Raptor: Robust and perception-aware trajectory
  replanning for quadrotor fast flight}.
\newblock {\it \bibinfo{journal}{IEEE Transactions on Robotics}\/},  {\it
  \bibinfo{volume}{37}\/}, \bibinfo{pages}{1992--2009}.

\end{thebibliography}

\end{document}